%% file: colm2026_conference.tex
\documentclass{article} %
\usepackage[preprint]{colm2026_conference}

\usepackage{microtype}
\usepackage{hyperref}
\usepackage{url}
\usepackage{booktabs}

\usepackage{graphicx}
\usepackage{enumitem}
\usepackage{multirow}
\usepackage{makecell}
\usepackage{pifont}
\usepackage{xspace}
\newcommand{\ours}{\textsc{EvoLM}\xspace}
\usepackage{amsmath}
\usepackage{wrapfig}
\usepackage[font=small,labelfont=bf]{caption}
\usepackage{subcaption}
\usepackage{fvextra}
\usepackage{float} %

\definecolor{forestgreen}{rgb}{0.13, 0.55, 0.13}
\newcommand{\cmark}{\textcolor{forestgreen}{\ding{51}}}
\newcommand{\xmark}{\textcolor{red}{\ding{55}}}

\usepackage{lineno}

\definecolor{darkblue}{rgb}{0, 0, 0.5}
\hypersetup{colorlinks=true, citecolor=darkblue, linkcolor=darkblue, urlcolor=darkblue}

\definecolor{discussionpurple}{rgb}{0.5, 0.0, 0.5}

\title{\ours: Self-Evolving Language Models\\through Co-Evolved Discriminative Rubrics}

\author{%
Shuyue Stella Li$^{1}$\footnotemark[1],\;\;
Rui Xin$^{1}$\footnotemark[1],\;\;
Teng Xiao$^2$,\;\;
Yike Wang$^1$,\;\;
Rulin Shao$^1$,\;\;
Zoey Hao$^3$\;\\
\textbf{Melanie Sclar$^{1}$},\;\;
\textbf{Sewoong Oh$^{1}$},\;\;
\textbf{Faeze Brahman$^{2}$},\;\;
\textbf{Pang Wei Koh$^{1,2}$},\;\;
\textbf{Yulia Tsvetkov$^1$} \\
$^1$University of Washington, $^2$Allen Institute for AI, $^3$University of Pennsylvania \\
\texttt{\{stelli,rx31\}@cs.washington.edu} \\
\parbox{0.03\textwidth}{\includegraphics[width=\linewidth]{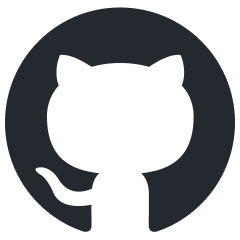}}\hspace{0.5mm}\href{https://github.com/stellalisy/EvoLM}{\texttt{https://github.com/stellalisy/EvoLM}}\\
\parbox{0.033\textwidth}{\includegraphics[width=\linewidth]{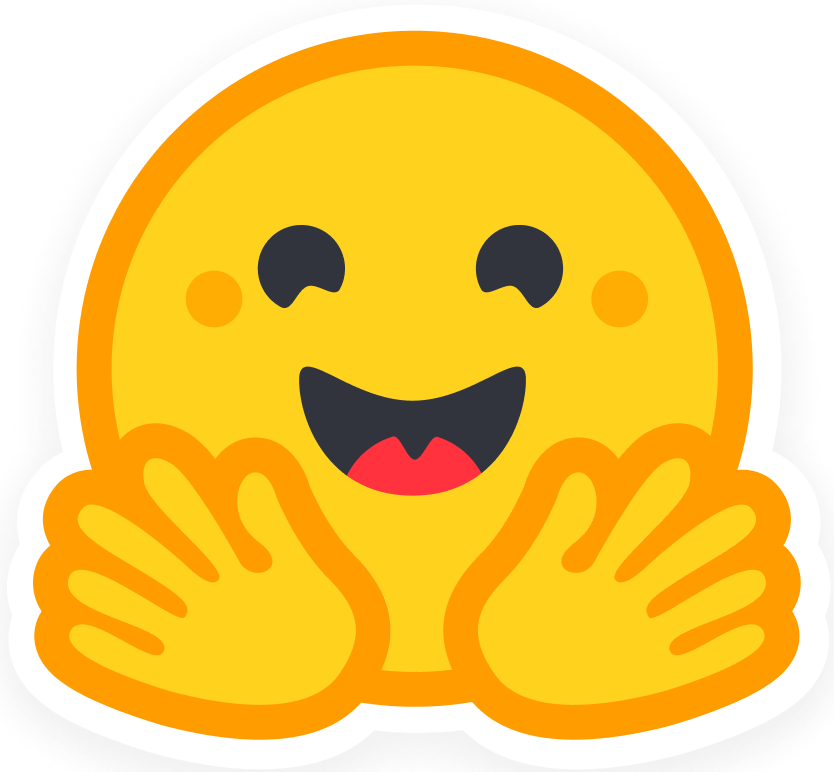}}\hspace{0.5mm}\href{https://huggingface.co/stellalisy/EvoLM-8B}{\texttt{https://huggingface.co/stellalisy/EvoLM-8B}}\vspace{-3mm}
}

\begin{document}

\ifcolmsubmission
\linenumbers
\fi

\maketitle

\begin{abstract}
Language models encode substantial evaluative knowledge from pretraining, yet current post-training methods rely on external supervision (human annotations, proprietary models, or scalar reward models) to produce reward signals. Each imposes a ceiling. Human judgment cannot supervise capabilities beyond its own, proprietary APIs create dependencies, and verifiable rewards cover only domains with ground-truth answers. Self-improvement from a model's own evaluative capacity is a reward source that scales with the model itself, yet remains largely untapped by current methods.
We introduce \ours, a post-training method that structures this capacity into explicit discriminative rubrics and uses them as training signal.
\ours trains two capabilities within a single language model in alternation:
(1) a rubric generator producing instance-specific evaluation criteria optimized for \emph{discriminative utility}, which maximizes a small frozen judge's ability to distinguish preferred from dispreferred responses; and (2) a policy trained using those rubric-conditioned scores as reward.
All preference signals are constructed from the policy's own outputs via temporal contrast with earlier checkpoints, requiring no human annotation or external supervision.
\ours trains a Qwen3-8B model to generate rubrics that outperform GPT-4.1 on RewardBench-2 by 25.7\%.
The co-trained policy achieves 69.3\% average on the OLMo3-Adapt suite, outperforming policies trained with GPT-4.1 prompted rubrics by 3.9\% and with the state-of-the-art 8B reward model SkyWork-RM by 16\%.
Overall, \ours
demonstrates that structuring a model's evaluative capacity into co-evolving discriminative rubrics enables self-improvement without external supervision.

\end{abstract}

\begin{figure}[h]
    \centering\vspace{-3mm}
    \includegraphics[width=\linewidth]{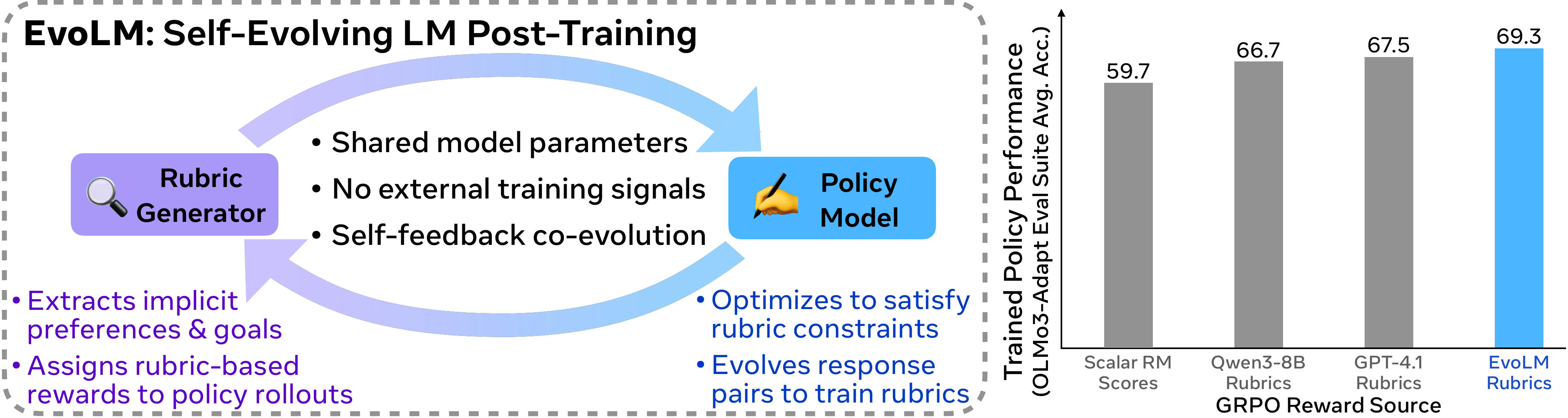}
    \caption{\ours Overview. Our method enables a single language model to co-evolve its own evaluation and generation capabilities. By extracting latent knowledge into explicit, instance-specific rubrics, EvoLM creates an autonomous feedback loop that improves policy performance without requiring human annotations or external teacher models.}
    \label{fig:teaser}%
\end{figure}

\section{Introduction}\vspace{-2mm}

Reinforcement learning (RL) has become central to language model (LM) post-training, driving improvements in instruction following, reasoning, and safety~\citep{ouyang2022training, bai2022constitutional}.
The quality of the reward signals fundamentally constrains what the policy can learn.
Current approaches derive this signal from external supervision, including human preference annotations~\citep{christiano2017deep}, scalar reward models susceptible to reward hacking~\citep{stiennon2020learning, gao2023scaling}, verifiable rewards restricted to domains with ground-truth answers~\citep{lambert2024tulu,cobbe2021training}, and proprietary LLM-as-a-judge evaluation~\citep{zheng2023judging}. %
Yet LMs already encode substantial evaluative knowledge from pretraining~\citep{feng2025dontthrowawaypretrained}; RL surfaces and sharpens this knowledge rather than introducing it~\citep{shao2026spuriousrewards}. Structuring this knowledge into a form that enables reliable scoring remains the central challenge for self-improvement.

In non-verifiable domains, evaluation can be factored into rubric generation (specifying what to measure) and judging (scoring against those criteria). This decomposition externalizes the model's evaluative knowledge into structured criteria, allowing even a small judge to score reliably when given concrete criteria that it could not derive on its own.
It also provides interpretability, where the rubric makes evaluation criteria inspectable, and modularity, such that a rubric generator trained with one judge can be deployed with another.
A growing body of work applies rubrics to RL post-training, but existing methods rely on external supervision such as task-specific verifiers or labeled preference data~\citep{sheng2026rlcer, xu2026rubricarm, lv2026learning}, or a proprietary teacher model for rubric generation~\citep{kim2023prometheus, gunjal2025rar, shen2026rrd, shao2025dr}. %
But to our knowledge, no existing method formally defines what makes a rubric useful or trains the rubric generator end-to-end to optimize for it without external supervision.

\begin{figure}
    \centering
    \includegraphics[width=\linewidth,height=6.5cm]{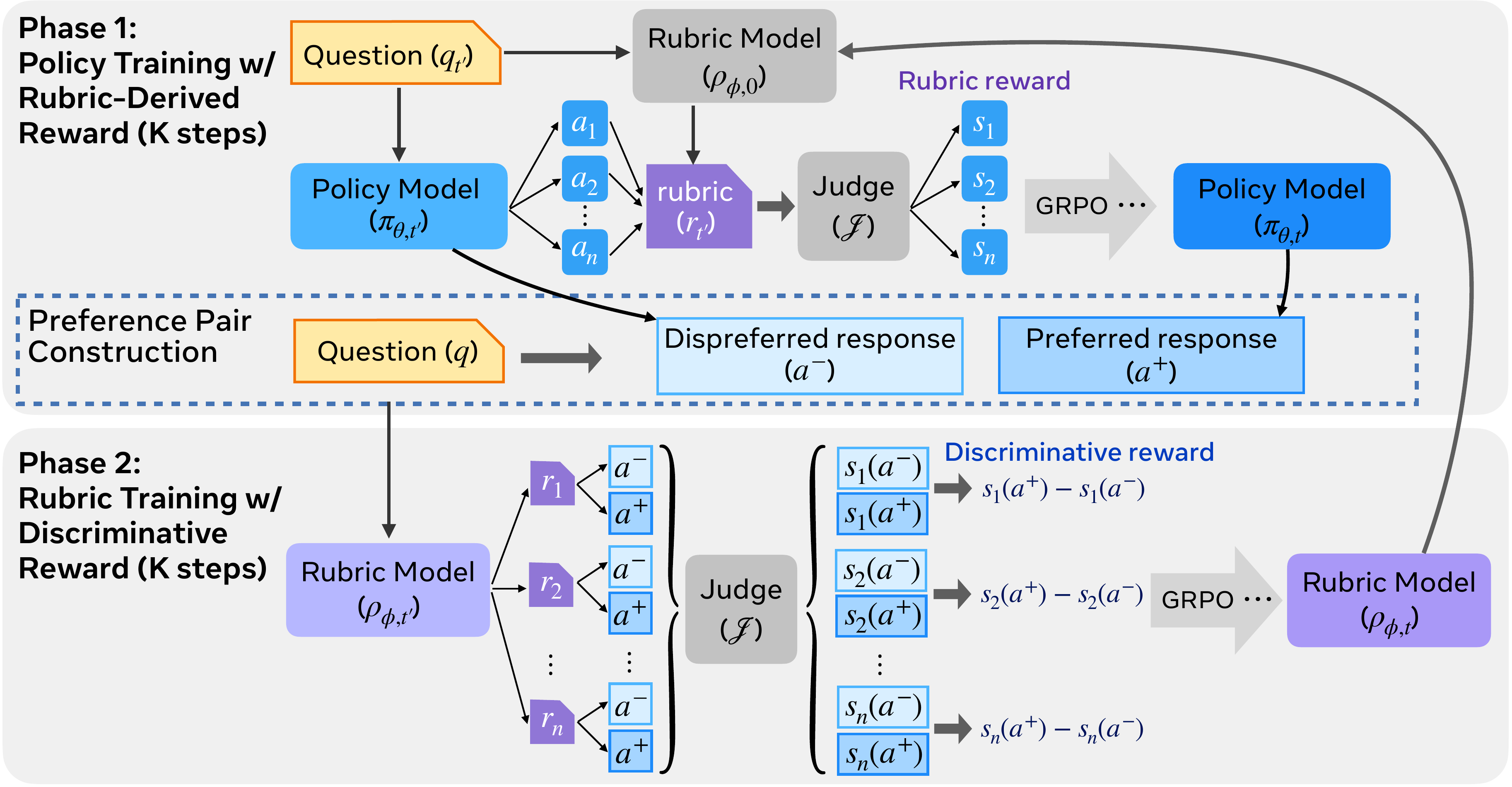}\vspace{-2mm}
    \caption{\ours alternates between two $K$-step phases. In Phase 1, a frozen rubric model $\rho_\phi$ generates a rubric $r$ for each question; the policy $\pi_\theta$ samples $n$ responses, which a frozen judge $\mathcal{J}$ scores to produce rubric-derived rewards used for policy updates. In Phase 2, the updated policy constructs preference pairs $(a^+, a^-)$ from its own outputs; the rubric model samples $n$ candidate rubrics, each scored by the judge on both responses, and the score margin $s(a^+) - s(a^-)$ serves as discriminative reward for rubric updates. The judge remains frozen throughout.}
    \label{fig:overview}\vspace{-3mm}
\end{figure}

This gap motivates a natural criterion for rubric quality: a rubric is useful when it helps the judge distinguish good responses from bad ones.
Given a preference pair, we can directly test whether the judge, conditioned on a candidate rubric, assigns a higher score to the preferred response.
Rubric quality is therefore directly measurable and trainable. We can optimize the rubric generator to maximize this \emph{discriminative utility}, rewarding rubrics that widen the judge's score gap between preferred and dispreferred responses.
The resulting training signal requires no human-written rubric annotations, no external labels, and no domain-specific verifiers; only preference pairs derived from the policy's own outputs.
These pairs are constructed via temporal contrast, where current policy responses (preferred) are paired against responses from earlier checkpoints (dispreferred).
We propose \textbf{\ours}, a post-training method in which a single language model learns to generate both responses and the evaluation rubrics used to score them.
The rubric generator and policy co-evolve through alternating updates (Figure~\ref{fig:overview}).
As the policy improves, rubrics must sharpen to remain discriminative; sharper rubrics in turn yield more informative reward, driving further policy improvement.
We formalize rubric generation as variational inference over latent rubric variables and derive a principled training objective (Section~\ref{sec:method}).

Across twelve benchmarks in math, code, general reasoning, knowledge, instruction following, and open-ended generation, \ours produces the strongest downstream policy (69.3\% average on the OLMo3-Adapt suite), outperforming both GPT-4.1-prompted rubrics (66.7\%) and four recent rubric-based RL methods that depend on proprietary APIs or labeled preference data (Section~\ref{sec:main}).
A scalar reward model that dominates held-out preference benchmarks (86.4\% RewardBench~2, 80.8\% JudgeBench) produces the weakest downstream policy (59.7\%). This 9.6-point gap behind \ours is consistent with prior findings on reward overoptimization ~\citep{gao2023scaling,ivison2024unpacking}. %
The self-evolving paradigm and the learned rubrics generalize broadly. The co-evolving framework extends to other model families including OLMo-3-7B, the learned rubrics transfer to unseen policies and judges without retraining, and on out-of-distribution deep research tasks, \ours achieves higher pairwise agreement with expert human rubrics than GPT-4.1 on both HealthBench (58.4\% vs.\ 52.5\%) and ResearchQA (59.3\% vs.\ 51.0\%) (\S\ref{sec:generalization}).
We further uncover the underlying self-evolving mechanism through qualitative analysis and show the trained generator learns to decompose evaluation into individually checkable sub-tasks, embedding expected intermediate values directly into criteria and transforming holistic judgment into pattern matching that even a 1.7B-parameter judge performs reliably.

\section{\ours}
\label{sec:method}%
\subsection{Overview}\vspace{-1.5mm}

Given a question $q$, a \textbf{policy} $\pi_\theta$ generates responses $a \sim \pi_\theta(\cdot \mid q)$. Standard approaches train a scalar reward model $R(q, a) \in \mathbb{R}$ that scores each response; the evaluation criteria are implicit in model weights and cannot be inspected. We instead decompose reward computation into two components with an explicit natural-language interface. A \textbf{rubric generator} $\rho_\phi$ produces evaluation criteria $r$, expressed in natural language, for each question:\vspace{-5mm}

\begin{equation}
r = \rho_\phi(q).
\end{equation}

\vspace{-2mm}
A \textbf{judge} $J$ scores responses against those criteria:\vspace{-4.5mm}

\begin{equation}
s = J(q, r, a) \in [0, 1].
\end{equation}
\vspace{-4mm}

We train the rubric generator and policy jointly, alternating between (a) policy updates using rubric-based judge scores as reward and (b) rubric generator updates using ranking accuracy of preference pairs from the policy's outputs as reward, with the judge's parameters fixed throughout (Figure~\ref{fig:overview}). The rest of this section describes in detail the training objective (\S\ref{sec:training_objective}), joint training procedure (\S\ref{sec:joint_training}), and preference pair construction (\S\ref{sec:preference_pairs}).

\subsection{Rubrics as Latent Variables} \label{sec:training_objective}\vspace{-1.5mm}

A rubric is good if it \textit{discriminates}: when the judge applies it, preferred responses score higher.
We can test whether a candidate rubric leads the judge to recover that ordering.
This motivates treating the rubric as a latent variable that explains observed preferences.
We formalize this through variational inference, deriving a training objective that rewards rubrics for reconstructing the preference ordering (Eq.~\ref{eq:elbo}).

For each observed preference pair, we introduce a latent variable $r \sim \rho_{\text{ref}}(r\mid q)$, representing the prior over rubrics from the base model. The likelihood of the observed preference given a latent rubric is:
\begin{align}
    p(a^+ \succ a^- \mid q, r) = \sigma\big( J(q, r, a^+) - J(q, r, a^-) \big),
\end{align}
where $\sigma$ is the sigmoid function. The joint distribution over responses, preferences, and rubrics is:
\begin{align}
    p(a^+, a^-, r \mid q)=  p(r\mid q)\, \pi(a^+|q)\, \pi(a^-|q)\, p(a^+ \succ a^- \mid q, r). \label{Eq:generation}
\end{align}
Here $\pi$ denotes the distribution from which responses are drawn; it can be a policy or an offline preference dataset. In our setting, $\pi = \pi_\theta$ is the current policy, and responses are held fixed when optimizing the rubric generator parameters $\phi$ (Section~\ref{sec:joint_training}).
Eq.~\ref{Eq:generation} generalizes the Bradley--Terry model~\citep{bradley1952rank}, which generates preferences from a latent scalar reward~\citep{rafailov2023direct}. We instead condition preference generation on a latent rubric variable, using natural language criteria rather than a single score.

We seek the posterior $p(r\mid a^+, a^-, q)$, which is intractable. Applying amortized variational inference~\citep{kingma2013auto}, we maximize the evidence lower bound (ELBO):
\begin{equation}
    \log p(a^+, a^- \mid q) \geq \mathcal{L}= \mathbb{E}_{r \sim \rho_{\phi}(r \mid q)}\big[\log p(a^+ \succ a^- \mid q, r)\big]-\text{KL}\big(\rho_{\phi}(r \mid q)\|\rho_{\text{ref}}(r \mid q) \big),\label{eq:elbo}
\end{equation}
where $\rho_{\phi}(r \mid q)$ is the variational distribution. Maximizing the ELBO brings $\rho_{\phi}$ close to the true posterior.
Because rubrics are discrete text, we optimize this objective with policy gradient, treating
the rubric generator acts as an agent that receives the log-likelihood of correct preference reconstruction as reward:
\begin{align}
   \mathcal{L}_{\text{rubric}}(\phi)= \mathbb{E}_{r \sim \rho_{\phi}(r \mid q)}\big[R(r; q, a^+, a^-)\big]-\text{KL}\big(\rho_{\phi}(r \mid q)\|\rho_{\text{ref}}(r \mid q) \big), \label{Eq:rubric-RL}
\end{align}
where $R(r; q, a^+, a^-)=\log \sigma\big( J(q, r, a^+) - J(q, r, a^-)\big)$ encourages rubrics that, under a fixed judge, assign higher scores to preferred responses.

\paragraph{Practical reward design.} The log-sigmoid reward in Equation~(\ref{Eq:rubric-RL}) provides a principled starting point. In practice, we replace it with a margin-based reward combined with a format reward:
\begin{equation}
    R(r; q, a^+, a^-) = \alpha \cdot \big(J(q, r, a^+) - J(q, r, a^-)\big) + (1 - \alpha) \cdot R_{\text{format}}(r)
\end{equation}
where $\alpha = 0.7$. The margin provides continuous signal proportional to the score gap. The format reward $R_{\text{format}}(r) \in \{0, 1\}$ verifies that the rubric conforms to a structured JSON schema the judge can parse (Appendix~\ref{app:format_reward}).
This reward design is ablated in Appendix~\ref{app:reward_shaping}.

\subsection{Joint Training} \label{sec:joint_training}

Both the rubric generator and the policy are trained using GRPO \citep{shao2024deepseekmath}, a policy gradient method that computes advantages from group-relative rewards without a learned value function. The policy's loss is:
\begin{equation}
    \mathcal{L}_{\text{policy}}(\theta) = -\mathbb{E}_{q, a \sim \pi_\theta}\left[A(q,a) \log \pi_\theta(a|q)\right]
\end{equation}
where advantages $A(q,a)$ are derived from judge scores $J(q, \rho_\phi(q), a)$ normalized within each group of responses to the same question. The rubric generator's loss $\mathcal{L}_{\text{rubric}}(\phi)$ (Equation~\ref{Eq:rubric-RL}) is optimized analogously, with advantages computed over groups of rubrics sampled for the same query.

Training alternates between two phases: the policy $\pi_\theta$ is updated for $K$ steps while the rubric generator $\rho_\phi$ is held fixed, then $\rho_\phi$ is updated for $K$ steps using preference pairs constructed from $\pi_\theta$'s outputs (Section~\ref{sec:preference_pairs}). This alternating structure creates an emergent curriculum.
The two phases reinforce each other. As the policy evolves, the rubric generator must learn increasingly specific criteria to distinguish quality under the current output distribution. As rubrics become more discriminative, the policy receives sharper reward signal and improves further.
The alternating frequency $K$ controls coupling strength and is ablated in Section~\ref{sec:ablations}.
We freeze judge $J$ throughout training, isolating the rubric generator as the sole source of improvement in training signal.

\subsection{Constructing Preference Pairs} \label{sec:preference_pairs}

The rubric generator's objective (Eq.~\ref{Eq:rubric-RL}) requires preference pairs $(a^+, a^-)$. We construct these entirely from the policy's own outputs without human annotation, using three complementary methods. In the default configuration, all three are active and sampled uniformly at random during rubric generator training. Each method is ablated independently in \S\ref{sec:ablations}.

\textbf{1. Temporal contrast.} We store policy rollouts with their training step. To construct a pair for question $q$, we generate a current response $a^+ \sim \pi_\theta(\cdot \mid q)$ at step $t$ and sample an earlier response $a^-$ from step $t' < t$. Since $\pi_\theta$ improves over training, the later response is treated as preferred. The step gap $t - t'$ controls the difficulty of discrimination: a large gap produces easy pairs with clear quality differences, while a small gap produces harder pairs that require more fine-grained rubrics. This creates a natural curriculum as training progresses, since the earliest responses are gradually replaced by stronger ones.

\textbf{2. Inferred question (IQ).} Given a preferred response $a^+$ to question $q$, the policy itself infers what question $a^+$ appears to address, producing $\hat{q}$, following~\citet{wang2026small}. A dispreferred response is then generated as $a^- \sim \pi_\theta(\cdot|\hat{q})$. This provides signal about whether the rubric captures question-relevance: rubrics that check whether the response addresses the intended question will score $a^+$ higher than $a^-$.

\textbf{3. Rubric-conditioned (RC).} Given question $q$, we sample a rubric $r \sim \rho_\phi(\cdot \mid q)$ from the current rubric generator and generate a preferred response $a^+ \sim \pi_\theta(\cdot \mid q, r)$ conditioned on both the question and rubric, and a dispreferred response $a^- \sim \pi_\theta(\cdot \mid q)$ conditioned on the question alone. This provides direct signal about rubric utility: if rubric-conditioned responses are consistently better, the rubric contains actionable criteria. The rubric generator is thus incentivized to produce criteria that concretely improve response quality when provided to the policy. Full prompts for all three methods are in Appendix~\ref{app:preference_prompts}.

\section{Experiments}

Experiments are structured around three questions: (1) Are trained rubric generators better than prompting, and does co-evolving training improve over sequential training (Section~\ref{sec:main})? (2) Do learned rubrics generalize across domains, policy architectures, and judges (Section~\ref{sec:generalization})? (3) What design choices most affect downstream policy quality (Section~\ref{sec:ablations})? %

\subsection{Setup}\label{sec:setup}

\textbf{Data.} Prompts are drawn from the Tulu 3 preference mixture
\citep{lambert2024tulu} and deduplicated, comprising approximately 271K prompts across
general-purpose chat (UltraFeedback, WildChat), instruction following
with IFEval-style constraints, math, code,
scientific literature understanding, and persona-driven synthetic
instructions. %

\textbf{Models.} Qwen3-8B \citep{qwen3} is used as both policy and rubric generator in a parameter-sharing configuration, where a single model handles both roles via different prompts. This minimizes memory requirements; a two-model configuration is ablated in Section~\ref{sec:ablations}. Qwen3-1.7B serves as the default frozen judge. The self-evolving framing predicts that even small judges should provide effective reward signal when given discriminative rubrics; using a 1.7B judge tests this directly. The judge size is ablated in Section~\ref{sec:ablations}.

\textbf{Training.} GRPO is used with learning rate $10^{-6}$, KL coefficient $0.001$, and 8 response samples per question. The alternating frequency $K=50$ controls how many steps each component trains before switching; this is ablated in Section~\ref{sec:ablations}. The temporal contrast step gap $[20, 100]$ controls the temporal distance between paired responses; this is also ablated in Section~\ref{sec:ablations}. Other implementation details are in Appendix~\ref{app:implementation}.

\textbf{Baselines.} We compare against three baseline families.

\textit{1. Prompted rubrics.} Two training-free baselines isolate the effect of learning the rubric generator. \textit{GPT-4.1 prompted} generates rubrics with GPT-4.1 and scores them with the same Qwen3-1.7B judge as \ours, representing a strong upper bound for prompting. \textit{Qwen3-8B prompted} matches our rubric-generator architecture but uses prompting rather than training.

\textit{2. Scalar reward model.} An alternative to natural language rubrics is a scalar reward model that encodes evaluation criteria implicitly in its weights. We use Skywork-RM-V2~\citep{liu2026skyworkrewardv2}, a state-of-the-art general-purpose Bradley-Terry reward model, to assign rewards to rollouts during GRPO. This baseline isolates whether rubric factorization provides value beyond a strong scalar reward signal.

\textit{3. Rubric-based RL methods.} We compare against four recent methods that also use rubrics during RL training.
\textbf{\textsc{RaR}}~\citep{gunjal2025rar} uses GPT-4.1 to generate rubrics (we replace the original proprietary judge with our Qwen3-1.7B judge for controlled comparison).
\textbf{\textsc{RRD}}~\citep{shen2026rrd} has GPT-4.1 iteratively refine rubrics and combines per-item scores with whitened-uniform weighting.
Both methods rely on a proprietary API for rubric generation and train only the policy; our method replaces this dependency with a trained local rubric generator.
\textbf{\textsc{RLCER}}~\citep{sheng2026rlcer} co-trains the rubric generator and policy but requires a task verifier for correctness filtering, limiting it to domains with ground-truth answers.
\textbf{\textsc{Rubric-ARM}}~\citep{xu2026rubricarm} co-trains the rubric generator and the judge using reference pairs, restricting it to tasks with ground-truth preference labels at training time.
Our method requires neither verifiers nor preference annotations during policy training.

To ensure a fair comparison, all methods are trained for 500 policy update steps. However, per-step compute varies: \textsc{RRD} and \textsc{RLCER} incur significantly more overhead from recursive refinement and per-rollout rubric generation, respectively. A detailed per-step cost breakdown is in Appendix~\ref{app:compute_accounting}.
Table~\ref{tab:baseline_design} highlights the key design-space differences. Appendix~\ref{app:baseline_configs} provides full implementation details.

\begin{table}[t]
\centering
\renewcommand\arraystretch{1.05}
\caption{Design-space comparison of rubric-based methods. Our method is the only one that trains the rubric generator, co-evolves it with the policy, and requires no proprietary API or external labels.}
\label{tab:baseline_design}\vspace{-3mm}
\small
\resizebox{\linewidth}{!}{
\begin{tabular}{lccccc}
\toprule
\textbf{Method} & \makecell{\textbf{Trains rubric}\\\textbf{generator}} & \makecell{\textbf{No proprietary}\\\textbf{API}} & \makecell{\textbf{No external}\\\textbf{labels}} & \makecell{\textbf{Not restricted to}\\\textbf{verifiable domains}} & \makecell{\textbf{Co-evolves}\\\textbf{with policy}} \\
\midrule
\textsc{RaR} & \xmark & \xmark & \cmark & \cmark& \xmark \\
\textsc{RRD} & \xmark & \xmark & \cmark & \cmark& \xmark \\
\textsc{RLCER} & \cmark & \cmark & \xmark & \xmark& \cmark \\
\textsc{Rubric-ARM} & \cmark & \cmark & \xmark & \cmark & \xmark \\
\textsc{\ours (Ours)} & \cmark & \cmark & \cmark & \cmark & \cmark \\
\bottomrule
\end{tabular}\vspace{-3mm}
}\vspace{-3mm}
\end{table}

\textbf{Evaluation.} We evaluate along two axes. First, \textit{downstream policy quality}: policies trained with learned rubrics are evaluated on twelve benchmarks from the OLMo3-Adapt suite, covering math reasoning (GSM8K, MATH), code generation (HumanEval+, MBPP+), general reasoning (BBH, GPQA, ZebraLogic, AGI-Eval), knowledge (MMLU, PopQA), instruction following (IFEval), and open-ended generation (AlpacaEval v3). Second, \textit{rubric quality}: we directly measure the discriminative accuracy of learned rubrics using RewardBench 2 \citep{malik2025rewardbench} and JudgeBench \citep{tan2025judgebenchbenchmarkevaluatingllmbased}, which test whether rubric-based evaluation correctly ranks preferred over dispreferred responses across diverse domains.

\subsection{Main Comparisons}
\label{sec:main}
\label{sec:main_comparisons}

In addition to the baselines described above, we compare two training regimes for our method.
\textbf{(1) Sequential training:} The rubric generator is trained first with the policy frozen, learning from preference pairs constructed via inferred question and rubric-conditioned methods (temporal contrast requires a changing policy). After rubric training converges, the policy is trained with the rubric generator frozen. This isolates the value of rubric training from the value of joint adaptation.
\textbf{(2) Co-evolving training (\ours):} The rubric generator and policy are trained jointly with alternating updates. This enables co-adaptation: as the policy improves, it generates more varied responses that provide richer preference signals; as the rubric generator improves, it provides sharper reward signals. All three preference signals methods including temporal contrast can be used.
We present our results in Table~\ref{tab:main}.

\begin{table*}[h]
\centering
\caption{Main comparison of training approaches. Each method produces a rubric generator (evaluated by RewardBench-2 and JudgeBench discriminative accuracy) and a policy (evaluated on OLMo3-Adapt downstream benchmarks). HE+ = HumanEval+. All scores are percentages.}
\label{tab:main}\vspace{-3mm}
\resizebox{\linewidth}{!}{
\begin{tabular}{lccccccc|cccccc}
\toprule
 & \multicolumn{7}{c|}{RewardBench2} & \multicolumn{6}{c}{JudgeBench} \\
Method & Factuality & Precise IF & Math & Safety & Focus & Ties & Avg & Knowledge & Reasoning & Math & Coding & Overall & Avg \\
\midrule
GPT-4.1 (prompted) & 32.1 & 32.6 & 34.4 & 39.1 & 35.0 & 46.3 & 36.6 & 41.6 & 48.0 & 30.4 & 45.2 & 42.0 & 41.3 \\
Qwen3-8B (prompted) & 27.8 & 26.5 & 38.1 & 25.0 & 30.2 & 9.7 & 26.2 & 34.4 & 38.8 & 30.4 & 42.9 & 36.0 & 36.6 \\
Scalar RM (Skywork-RM-V2) & \textbf{88.2} & \textbf{67.8} & \textbf{83.1} & \textbf{97.3} & \textbf{99.2} & \textbf{83.1} & \textbf{86.4} & \textbf{77.3} & \textbf{75.5} & \textbf{89.3} & \textbf{81.0} & \textbf{79.1} & \textbf{80.8} \\
\midrule
\textsc{RaR} & \multicolumn{7}{c|}{N/A (frozen GPT-4.1 rubric)} & \multicolumn{6}{c}{N/A (frozen GPT-4.1 rubric)} \\
\textsc{RRD}  & \multicolumn{7}{c|}{N/A (frozen GPT-4.1 rubric)} & \multicolumn{6}{c}{N/A (frozen GPT-4.1 rubric)} \\ %
\textsc{RLCER } & 33.1 & 29.8 & 52.2 & 48.3 & 51.9 & 54.8 & 45.0 & 43.5 & 40.8 & 48.2 & 64.3 & 46.0 & 49.2 \\ %
\textsc{Rubric-ARM}\footnote{Rubric-ARM uses a pairwise judge (comparing both responses jointly) while our evaluation scores each response independently against the rubric. Additionally, Rubric-ARM uses greedy decoding (temperature 0) for rubric generation. Our reproduction obtains 39.1\% Precise IF and 66.1\% Focus, comparable to Rubric-ARM's reported 41.9\% and 69.8\% on these subsets.} & 48.8 & 39.1 & 63.5 & 60.8 & 66.1 & 58.2 & 56.1 & 43.5 & 42.9 & 62.5 & 57.1 & 48.0 & 51.5 \\
\midrule
Sequential (IQ) & 37.0 & 32.0 & 57.1 & 42.3 & 59.0 & 55.6 & 47.2 & 37.7 & 55.1 & 44.6 & 57.1 & 46.0 & 48.6 \\
Sequential (RC) & 31.5 & 30.8 & 57.0 & 47.9 & 49.6 & 59.2 & 46.0 & 34.4 & 39.8 & 44.6 & 50.0 & 39.4 & 42.2 \\
\ours & 37.2 & 30.6 & 54.0 & 43.5 & 54.4 & 56.1 & 46.0 & 45.5 & 39.8 & 30.4 & 61.9 & 43.4 & 44.4 \\
\bottomrule
\end{tabular}
}
\vspace{0.5em}
\resizebox{\linewidth}{!}{
\begin{tabular}{lccccccccccccc}
\toprule
\multicolumn{14}{c}{Downstream Policy Quality (OLMo3-Adapt)} \\
Method & GSM8K & MATH & HE+ & MBPP+ & BBH & MMLU & IFEval & PopQA & GPQA & Zebra & AGI-E & AlpacaE & Avg \\
\midrule
GPT-4.1 (prompted) & 95.7 & 94.2 & 58.7 & 65.1 & \textbf{71.7} & 84.8 & 34.4 & 30.7 & 54.9 & 81.9 & 82.9 & 45.5 & 66.7 \\
Qwen3-8B (prompted) & 95.5 & 94.6 & 73.7 & 58.4 & 66.8 & 83.4 & 35.3 & 31.7 & 56.9 & \textbf{84.3} & 83.9 & \textbf{45.9} & 67.5 \\
Scalar RM (Skywork-RM-V2) & 95.6 & 94.3 & 47.9 & 66.2 & 54.9 & 82.0 & 37.3 & 25.2 & 46.2 & 56.4 & 81.6 & 28.3 & 59.7 \\
\midrule
\textsc{RaR} & 95.7 & 93.8 & 77.4 & \textbf{68.6} & 62.5 & 84.8 & 36.0 & 32.2 & 52.5 & 84.1 & 87.3 & 34.7 & 67.5 \\
\textsc{RRD}  & 95.3 & 92.2 & 75.2 & 59.2 & 69.4 & 84.8 & 36.8 & 32.6 & 56.0 & 82.6 & \textbf{88.9} & 38.5 & 67.6 \\ %
\textsc{RLCER}  & 94.0 & \textbf{94.7} & 80.5 & 67.9 & 68.6 & 84.8 & \textbf{39.2} & 30.4 & 49.6 & 69.1 & 84.3 & 37.2 & 66.7 \\
\textsc{Rubric-ARM} & 95.6 & 93.3 & 78.4 & 62.3 & 64.4 & 83.8 & 35.7 & \textbf{32.8} & 56.9 & 83.1 & 87.2 & 35.9 & 67.5 \\
\midrule
Sequential (IQ) & 95.9 & 92.7 & 76.5 & 66.7 & 64.6 & 85.0 & 38.1 & 32.0 & 54.0 & 83.0 & 86.9 & 41.1 & 68.0 \\
Sequential (RC) & \textbf{96.0} & 93.4 & 80.2 & 67.7 & 64.7 & \textbf{85.8} & 36.8 & 32.3 & \textbf{57.1} & 83.2 & 87.4 & 35.1 & 68.3 \\
\ours & 95.8 & 94.5 & \textbf{86.2} & 68.5 & 67.6 & 85.3 & 37.7 & 30.5 & 54.2 & 82.3 & 86.4 & 42.2 & \textbf{69.3} \\
\bottomrule
\end{tabular}\vspace{-3mm}
}
\end{table*}

The policy model trained by \ours achieves the highest downstream average (69.3\%), outperforming all baselines without proprietary APIs, external labels, or task-specific verifiers. Prompted rubrics from GPT-4.1 (66.7\%) and Qwen3-8B (67.5\%) are competitive on individual benchmarks but fall behind on the aggregate, as are the four rubric-based RL methods (66.7--67.6\%). The largest gains are on code generation (HumanEval+ 86.2\% vs.\ 80.5\% for the next best method), where the fine-grained criteria produced by trained rubrics provide particularly sharp signal. Co-evolving training outperforms sequential training on policy quality (69.3\% vs.\ 68.3\%), despite sequential training achieving higher static rubric accuracy (RB2 47.2\% vs.\ 46.0\%). Downstream policy quality depends on how well rubrics adapt to the evolving policy distribution, precisely the advantage co-evolution provides.

The scalar reward model presents a striking contrast. It dominates RewardBench-2 (86.4\%) and JudgeBench (80.8\%) yet produces the weakest downstream policy (59.7\%), falling 9.6 points behind \ours. This is consistent with reward overoptimization~\citep{gao2023scaling, ivison2024unpacking}. Static evaluation criteria, whether encoded in model weights or in a fixed rubric, cannot adapt to the reward landscape that a learning policy creates. Co-evolving rubrics avoid this failure mode by continuously restructuring evaluation criteria to remain discriminative against the current policy distribution. We analyze the mechanism (how rubrics evolve from abstract labels to verifiable checks) in Section~\ref{sec:qualitative}.

\subsubsection{Rubrics Evolve from Abstract Labels to Verifiable Checks}\label{sec:qualitative}

Qualitative analysis across four domains (math, emotional support, scientific explanation, and constrained writing) reveals a pattern: early rubrics use short labels or generic checks, while trained rubrics pack specific, verifiable expectations into each criterion, reducing the interpretive burden on the judge model. The form this takes varies by domain.
In math, for instance, the rubric concentrates 80\% of the weight on a single criterion that embeds the expected answer (e.g., ``the correct maximum area of 144, derived from the given perimeter of 48''), transforming proof verification into answer checking; in constrained writing, it consolidates formatting and keyword requirements into explicit, countable checks. Aggregate statistics over the 100 evaluation prompts confirm this shift: label-only criteria fall from 21.9\% to 0.3\%, criteria embedding specific expected values rise from 6.9\% to 19.3\%, and constraint-type criteria rise from 7.7\% to 20.3\%, all relative to prompted Qwen3-8B (Appendix~\ref{app:rubric_stats}). The common effect is moving evaluation from holistic semantic judgment, which small judges perform unreliably, to pattern matching over concrete criteria (Appendix~\ref{app:rubric_examples}).
This is a direct consequence of the margin objective (Eq.~\ref{Eq:rubric-RL}) that any criterion that widens the judge's score gap between preferred and dispreferred responses is rewarded, which favors concrete, verifiable criteria over abstract ones.
This enrichment progresses over time. Criteria length grows monotonically (59 to 112 characters on average) while criteria count remains stable at ${\sim}3$--$4$, and in some domains rubrics briefly decompose constraints into many fine-grained items before reconsolidating them into fewer, denser criteria (Appendix~\ref{app:rubric_evolution}).

\subsection{Generalization}
\label{sec:generalization}

A rubric generator has practical value only if it generalizes beyond the training distribution. The self-evolving process optimizes rubrics for a specific policy and judge; we now test whether it nonetheless captures transferable evaluation principles across unseen domains, policy architectures, and judges.

\subsubsection{Alignment to Expert Deep Research Rubrics}\label{sec:dr}

\ours is trained on the Tulu~3 dataset \citep{lambert2024tulu}, which contains general-purpose tasks (\S\ref{sec:setup}). Deep research tasks that require long-form, multi-step responses are entirely unseen during rubric generator training. We test whether the learned rubric generator nonetheless produces evaluation criteria that align with expert human rubrics on these out-of-distribution tasks.

We evaluate on HealthBench \citep{arora2025healthbench} and ResearchQA \citep{yifei2025researchqa}, two benchmarks that provide expert-written rubrics for each question. For each question, we generate rollouts from DR Tulu-8B SFT and RL checkpoints \citep{shao2025dr} to ensure response quality diversity, then obtain ground-truth pairwise rankings by grading rollouts against the human expert rubrics using GPT-4.1. We generate rubrics using \ours, prompted Qwen3-8B, and prompted GPT-4.1, then score rollouts using Qwen3-1.7B to obtain pairwise rankings. We report pairwise ranking accuracy and Acc@$\delta$, restricted to pairs where the expert-rubric score gap exceeds threshold $\delta$ to filter near-ties where the ground-truth ranking is ambiguous.

\begin{wrapfigure}{R}{0.5\textwidth}\vspace{-3.8mm}
\setlength{\tabcolsep}{2pt}
\centering
\captionof{table}{Alignment to expert rubrics on OOD tasks.}
\label{tab:rubric-eval-standard}\vspace{-2mm}
\resizebox{\linewidth}{!}{
\begin{tabular}{lccccccc}
\toprule
& \multicolumn{2}{c}{\textbf{HealthBench}}  & \multicolumn{2}{c}{\textbf{ResearchQA}} \\
\cmidrule(lr){2-3} \cmidrule(lr){4-5}
\textbf{Rubric Generator} & \textbf{Acc} & \textbf{Acc@0.1} & \textbf{Acc} & \textbf{Acc@0.05} \\
\midrule
\ours                        & \textbf{58.4} & \textbf{59.0} & \textbf{59.3} & \textbf{68.7} \\
Qwen3-8B (prompted)          & 53.0 & 55.0 & 57.2 & 59.0 \\
GPT-4.1 (prompted)           & 52.5 & 53.6 & 51.0 & 65.3 \\
\bottomrule
\end{tabular}\vspace{-2mm}
}
\end{wrapfigure}

As shown in Table~\ref{tab:rubric-eval-standard}, \ours achieves the highest pairwise ranking accuracy on both benchmarks: 58.4\% on HealthBench and 59.3\% on ResearchQA, outperforming prompted Qwen3-8B (53.0\%, 57.2\%) and GPT-4.1 (52.5\%, 51.0\%) by 5--8 points. The advantage holds and widens on the filtered metrics, reaching 59.0\% Acc@0.1 on HealthBench and 68.7\% Acc@0.05 on ResearchQA. That a rubric generator trained on general-purpose tasks produces criteria better aligned with expert judgment than GPT-4.1 on health and research domains provides strong evidence that the variational training objective learns transferable evaluation structure. This strong OOD performance, despite the disconnect between static preference benchmarks and downstream rubric utility observed in \S\ref{sec:main_comparisons}, further supports evaluating rubric generators on their ability to align with expert judgment rather than on held-out preference accuracy alone.

\subsubsection{Co-evolved rubrics provide effective training signals to unseen policies}

To test whether learned rubrics encode general evaluation criteria rather than policy-specific ones, we freeze the rubric generator from the main experiment and use it to train two unseen policies from scratch: Qwen3-4B (same family, smaller) and Llama-3.1-8B (different family). Each policy is also trained with GPT-4.1-prompted rubrics as a controlled comparison.

\begin{table*}[h]
\centering
\caption{Cross-model transfer. The rubric generator (frozen from the main Qwen3-8B experiment) is used to train policies of different sizes and families. Downstream quality measures whether the rubric provides effective reward signal for models not seen during rubric training. Rubric quality (RewardBench-2/JudgeBench) is not reported because the rubric generator is frozen. All scores are percentages.}
\label{tab:cross_model}\vspace{-3mm}
\resizebox{\linewidth}{!}{
\begin{tabular}{lccccccccccccc}
\toprule
\multicolumn{14}{c}{Downstream Policy Quality (OLMo3-Adapt)} \\
Configuration & GSM8K & MATH & HE+ & MBPP+ & BBH & MMLU & IFEval & PopQA & GPQA & Zebra & AGI-E & AlpacaE & Avg \\
\midrule
Qwen3-4B + GPT-4.1 rubric & \textbf{94.8} & 91.9 & 70.1 & 67.7 & 61.9 & \textbf{82.8 }& \textbf{37.0} & \textbf{26.5 }& 45.5 & 79.7 & \textbf{84.6 }& \textbf{30.5} & 64.4 \\
Qwen3-4B + \ours rubric & 94.5 & \textbf{92.4 }& \textbf{77.9} & \textbf{68.0} & \textbf{65.3} & 78.8 & 36.8 & 25.5 & \textbf{49.3 }& \textbf{79.9} & 83.3 & 30.4 & \textbf{65.2} \\
\midrule
Llama-3.1-8B + GPT-4.1 rubric & 73.0 & 43.7 & \textbf{49.1 }& \textbf{42.3} & \textbf{56.8 }& 58.8 & 67.5 & 25.8 & 33.7 & 11.3 & \textbf{63.6} & \textbf{22.6} & 45.7 \\
Llama-3.1-8B + \ours rubric & \textbf{79.8} & \textbf{47.2} & 40.5 & 38.9 & 56.0 & \textbf{71.4} & \textbf{68.4} & \textbf{31.6} & \textbf{35.5 }& \textbf{12.0} & 63.4 & 18.4 & \textbf{46.9} \\
\bottomrule
\end{tabular}
}
\end{table*}

As shown in Table~\ref{tab:cross_model}, \ours rubrics outperform GPT-4.1-prompted rubrics on both architectures: Qwen3-4B (65.2\% vs.\ 64.4\%) and Llama-3.1-8B (46.9\% vs.\ 45.7\%). Within the Qwen3 family, the largest gains appear on code generation (HE+ 77.9\% vs.\ 70.1\%) and general reasoning (GPQA 49.3\% vs.\ 45.5\%, BBH 65.3\% vs.\ 61.9\%). These results confirm that the rubric generator captures transferable evaluation criteria that provide effective reward signal for models not seen during rubric training.

\subsubsection{Trained rubrics generalize across judge models}

We test whether trained rubrics remain effective when paired with judges not seen during training. We freeze the \ours rubric generator (trained with Qwen3-1.7B judge) and evaluate discriminative accuracy with three unseen inference-time judges: Qwen3-8B, OLMo-3-7B-Instruct, and Mistral-7B (Table~\ref{tab:cross_judge}).

\begin{table}[h]
\centering
\caption{Cross-judge evaluation. RewardBench-2 and JudgeBench accuracy with different inference-time judges. The rubric generator was trained with Qwen3-1.7B. Bold indicates the higher value within each prompted/trained pair.}
\label{tab:cross_judge}\vspace{-3mm}
\resizebox{\linewidth}{!}{
\begin{tabular}{lcccccccc|cccccc}
\toprule
\multirow{ 2}{*}{Inference Judge} & \multirow{ 2}{*}{Rubric} & \multicolumn{7}{c|}{RewardBench2} & \multicolumn{6}{c}{JudgeBench} \\
 && Factuality & Precise IF & Math & Safety & Focus & Ties & Avg & Knowledge & Reasoning & Math & Coding & Overall & Avg \\
\midrule
Qwen3-1.7B & Prompted & 27.8 & 26.5 & 38.1 & 25.0 & 30.2 & 9.7 & 26.2 & 34.4 & 38.8 & 30.4 & 42.9 & 36.0 & 36.6 \\
(training judge) & Trained & \textbf{37.2} & \textbf{30.6} & \textbf{54.0} & \textbf{43.5} & \textbf{54.4} & \textbf{56.1} & \textbf{46.0} & \textbf{45.5} & \textbf{39.8} & 30.4 & \textbf{61.9} & \textbf{43.4} & \textbf{44.4} \\
\addlinespace
\multirow{ 2}{*}{Qwen3-8B} & Prompted & 37.4 & 33.6 & 51.5 & 49.9 & 45.5 & 20.3 & 39.7 & 48.1 & 33.7 & 53.6 & 61.9 & 46.6 & 49.3 \\
 & Trained & \textbf{49.9} & \textbf{41.7} & \textbf{74.3} & \textbf{55.3} & \textbf{71.0} & \textbf{82.2} & \textbf{62.4} & \textbf{57.1} & \textbf{68.4} & \textbf{66.1} & 61.9 & \textbf{62.3} & \textbf{63.4} \\
\addlinespace
\multirow{ 2}{*}{OLMo-3-7B-Instruct} & Prompted & 36.2 & 24.7 & \textbf{57.7} & \textbf{48.7} & \textbf{43.2} & 32.5 & 40.5 & 40.9 & \textbf{32.7} & \textbf{37.5} & 40.5 & 38.0 & 37.9 \\
 & Trained & \textbf{37.9} & \textbf{26.4} & 54.0 & 48.0 & 39.9 & \textbf{57.6} & \textbf{43.9} & \textbf{42.9} & 31.6 & 26.8 & \textbf{59.5} & \textbf{39.1} & \textbf{40.2} \\
\addlinespace
\multirow{ 2}{*}{Mistral-7B} & Prompted & 27.8 & 22.9 & \textbf{28.6} & \textbf{38.5} & 28.5 & \textbf{34.2} & \textbf{30.1} & 31.2 & 17.3 & 28.6 & 11.9 & 24.6 & 22.2 \\
 & Trained & \textbf{29.4} & \textbf{25.4} & 28.5 & 31.9 & \textbf{35.0} & 22.1 & 28.7 & \textbf{36.4} & \textbf{34.7} & \textbf{33.9} & \textbf{38.1} & \textbf{35.7} & \textbf{35.8} \\
\bottomrule
\end{tabular}
}\vspace{-3mm}
\end{table}

The strongest result is on Qwen3-8B, where trained rubrics improve RewardBench-2 by $+22.7$ points (62.4\% vs.\ 39.7\%) and JudgeBench by $+15.7$ points over prompted rubrics. This indicates that trained rubrics encode evaluation structure that a more capable judge can leverage more effectively than it can derive from a generic prompt alone. Cross-family transfer is more modest but positive on aggregate: OLMo-3-7B-Instruct gains $+3.4$ on RewardBench-2 and Mistral-7B gains $+11.1$ on JudgeBench. Together, these results show that a rubric generator trained with one judge can be deployed with different judges without retraining.

\subsubsection{Multi-Judge Training}

To produce rubric generators that generalize out of the box to unseen judges, we train with multiple judges simultaneously. We use a five-judge ensemble of small models: Qwen3-1.7B, Llama-3.2-1B-Instruct, OLMo-2-0425-1B-Instruct, Gemma-3-1B-IT, and Qwen3-4B. Each judge $j \in \{1, \dots, J\}$ independently scores the accepted and rejected answers, $a^+$ and $a^-$, using the generated rubric, obtaining $\text{score}_j^+$ and $\text{score}_j^-$, and casts a binary vote $v_j = \mathbf{1}[\text{score}_j^+ > \text{score}_j^-]$. The reward for each rollout is
\begin{equation}
R_{\text{MJ}} = w_m \cdot \bar{m} + w_f \cdot \mathbf{1}[\text{valid\_format}] + w_\kappa \cdot \hat{\kappa},
\label{eq:mkf_reward}
\end{equation}
where $\bar{m} = \frac{1}{J}\sum_{j=1}^{J} (s_j^+ - s_j^-)$ is the average margin (difference between accepted and rejected scores across judges), $\mathbf{1}[\text{valid\_format}]$ is a binary indicator for whether the generated rubric satisfies the required JSON schema, and $\hat{\kappa}$ is the \textbf{agreement among the judges}, calculated as $\hat{\kappa} = \text{clip}(\kappa_F(\vec{v}), 0, 1)$, where $\kappa_F$ is Fleiss's kappa \citep{fleiss1971measuring} computed over the binary vote from the judges of whether the accepted answer received a higher score than the rejected answer.
We test two weight configurations: \textbf{Margin+Format} (MF) with $(w_m, w_f, w_\kappa) = (0.7, 0.3, 0)$, which omits the agreement penalty entirely, and \textbf{Margin+Agreement+Format} (MAF) with $(w_m, w_f, w_\kappa) = (0.5, 0.3, 0.2)$, which retains a moderate agreement incentive.
Additionally, we consider the alternative reward design of using the binary vote and the agreement (BA) without the margin and format rewards:
\begin{equation}
R = \bar{v} - (1 - \hat{\kappa}), \quad \bar{v} = \frac{1}{J}\sum_{j=1}^{J} v_j, \quad \hat{\kappa} = \text{clip}(\kappa_F, 0, 1),
\label{eq:multi_judge_reward}
\end{equation}
where $\bar{v}$ is the average vote (fraction of judges preferring the accepted answer).
All multi-judge variants maintain comparable downstream policy quality to the single-judge baseline (full ablation in Appendix~\ref{app:multi_judge}).

\begin{table*}[t]
\centering
\caption{Cross-judge evaluation of single-judge and multi-judge rubric generators. RewardBench-2 and JudgeBench accuracy (\%) with three inference-time judges. The single-judge generator (\ours) was trained with Qwen3-1.7B; multi-judge generators were trained with five judges (Qwen3-1.7B, Llama-3.2-1B-Instruct, OLMo-2-0425-1B-Instruct, Gemma-3-1B-IT, Qwen3-4B). Multi-judge uses Eq.~\ref{eq:multi_judge_reward}; MF and MAF use Eq.~\ref{eq:mkf_reward} with weights $(0.7,0.3,0)$ and $(0.5,0.3,0.2)$, respectively.}\vspace{-3mm}
\label{tab:cross_judge_mj}
\resizebox{\linewidth}{!}{
\begin{tabular}{llccccccc|cccccc}
\toprule
\multirow{2}{*}{Inference Judge} & \multirow{2}{*}{Rubric Generator} & \multicolumn{7}{c|}{RewardBench2} & \multicolumn{6}{c}{JudgeBench} \\
 && Factuality & Precise IF & Math & Safety & Focus & Ties & Avg & Knowledge & Reasoning & Math & Coding & Overall & Avg \\
\midrule
\multirow{5}{*}{Qwen3-1.7B} & Prompted & 27.8 & 26.5 & 38.1 & 25.0 & 30.2 & 9.7 & 26.2 & 34.4 & 38.8 & 30.4 & 42.9 & 36.0 & 36.6 \\
 & \ours (single judge) & 37.2 & \textbf{30.6} & 54.0 & 43.5 & \textbf{54.4} & 56.1 & 46.0 & \textbf{45.5} & 39.8 & 30.4 & \textbf{61.9} & 43.4 & 44.4 \\
 & MF (Eq.~\ref{eq:mkf_reward}) & \textbf{37.4} & 29.4 & \textbf{56.5} & \textbf{50.3} & 49.8 & 53.8 & \textbf{46.2} & \textbf{48.7} & 45.9 & 42.9 & 59.5 & \textbf{48.3} & 49.3 \\
 & MAF (Eq.~\ref{eq:mkf_reward}) & 33.4 & 28.4 & 53.4 & 46.8 & 50.9 & \textbf{57.7} & 45.1 & 42.9 & 42.9 & \textbf{44.6} & 52.4 & 44.3 & 45.7 \\
 & BA (Eq.~\ref{eq:multi_judge_reward}) & 35.5 & 30.0 & 55.4 & 47.1 & 50.1 & 56.2 & 45.7 & 45.5 & \textbf{49.0} & 42.9 & \textbf{64.3} & \textbf{48.3} & \textbf{50.4} \\
\addlinespace
\multirow{5}{*}{Qwen3-8B} & Prompted & 37.4 & 33.6 & 51.5 & 49.9 & 45.5 & 20.3 & 39.7 & 48.1 & 33.7 & 53.6 & 61.9 & 46.6 & 49.3 \\
 & \ours (single judge) & 49.9 & \textbf{41.7} & 74.3 & 55.3 & \textbf{71.0} & 82.2 & 62.4 & 57.1 & \textbf{68.4} & \textbf{66.1} & 61.9 & \textbf{62.3} & \textbf{63.4} \\
 & MF (Eq.~\ref{eq:mkf_reward}) & 51.0 & 36.0 & 74.2 & \textbf{59.8} & 69.7 & 76.5 & 61.2 & 57.1 & 62.2 & 64.3 & 66.7 & 60.9 & 62.6 \\
 & MAF (Eq.~\ref{eq:mkf_reward}) & 55.9 & 35.5 & 77.6 & 56.2 & 70.5 & 75.8 & 61.9 & \textbf{57.8} & \textbf{71.4} & \textbf{66.1} & \textbf{69.0} & \textbf{64.3} & \textbf{66.1} \\
 & BA (Eq.~\ref{eq:multi_judge_reward}) & \textbf{57.6} & 36.1 & \textbf{78.5} & 52.5 & 70.9 & \textbf{85.6} & \textbf{63.5} & 57.1 & 69.4 & 64.3 & 61.9 & 62.3 & 63.2 \\
\addlinespace
\multirow{5}{*}{OLMo-3-7B} & Prompted & 36.2 & 24.7 & \textbf{57.7} & 48.7 & \textbf{43.2} & 32.5 & 40.5 & 40.9 & 32.7 & 37.5 & 40.5 & 38.0 & 37.9 \\
 & \ours (single judge) & 37.9 & 26.4 & 54.0 & 48.0 & 39.9 & \textbf{57.6} & 43.9 & \textbf{42.9} & 31.6 & 26.8 & \textbf{59.5} & \textbf{39.7} & \textbf{40.2} \\
 & MF (Eq.~\ref{eq:mkf_reward}) & 32.3 & 22.3 & 49.8 & 51.8 & 35.4 & 38.0 & 38.3 & 40.3 & 32.6 & \textbf{48.2} & 30.9 & 38.3 & 38.0 \\
 & MAF (Eq.~\ref{eq:mkf_reward}) & 33.0 & 25.7 & 54.7 & 50.3 & 35.7 & 37.6 & 39.5 & 39.0 & 29.6 & 25.0 & 30.9 & 33.1 & 31.1 \\
 & BA (Eq.~\ref{eq:multi_judge_reward}) & \textbf{38.7} & \textbf{27.6} & 56.3 & \textbf{52.7} & 40.3 & 54.7 & \textbf{45.1} & \textbf{53.2} & \textbf{41.8} & 37.5 & \textbf{52.4} & \textbf{47.4} & \textbf{46.2} \\
\bottomrule
\end{tabular}
}\vspace{-5mm}
\end{table*}

The key advantage of multi-judge training appears at inference time. Table~\ref{tab:cross_judge_mj} evaluates all multi-judge rubric generators against the single-judge generator (\ours) and prompted rubrics, using three inference-time judges not in the training mix: Qwen3-1.7B (the single-judge training judge), Qwen3-8B, and OLMo-3-7B-Instruct. All multi-judge generators consistently outperform both the single-judge generator and prompted rubrics across all three inference judges. On Qwen3-8B, the best multi-judge variant (MAF) achieves 61.9\% RewardBench-2 and 66.1\% JudgeBench, compared to 62.4\% and 63.4\% for the single-judge generator, a +2.7 point gain on JudgeBench. The improvement is even more pronounced on OLMo-3-7B-Instruct, where BA raises JudgeBench from 40.2\% to 46.2\% (+6.0). On the training judge Qwen3-1.7B, all multi-judge variants improve JudgeBench substantially (45.7--50.4\% vs.\ 44.4\%). These results suggest that training with diverse judges produces rubrics that are more universally interpretable, encoding evaluation criteria that transfer effectively to judges outside the training ensemble.

\begin{wrapfigure}{R}{0.45\textwidth}\vspace{-4.5mm}
\setlength{\tabcolsep}{2pt}
\centering%
\captionof{table}{Summary of ablation results. Each row varies one design dimension while holding others at the main configuration ($\star$). Rubric = RewardBench-2 average. Policy = OLMo3-Adapt 12-benchmark average (\%).}
\label{tab:ablation_summary}\vspace{-3mm}
\resizebox{\linewidth}{!}{
\begin{tabular}{llcc}
\toprule
\textbf{Dimension} & \textbf{Variant} & \textbf{RB2} & \textbf{Policy} \\
\midrule
\multirow{4}{*}{Reward design} & 5--10 criteria, no-dealbreaker prompt & 38.9 & 67.6 \\
& main prompt & 37.9 & \textbf{69.5} \\
& main prompt + margin & 37.2 & 66.9 \\
& main prompt + margin + format$^\star$ & \textbf{46.0} & 69.3 \\
\midrule
\multirow{5}{*}{Alt.\ freq.\ $K$} & 2 & 46.4 & 67.9 \\
& 10 & 44.5 & 68.5 \\
& 20 & 46.9 & 68.6 \\
& 50$^\star$ & 46.0 & \textbf{69.3} \\
& 100 & \textbf{48.5} & 68.7 \\
\midrule
\multirow{2}{*}{Model config.} & Single-model$^\star$ & 46.0 & \textbf{69.3} \\
& Two-model & \textbf{48.4} & \textbf{69.3} \\
\midrule
\multirow{5}{*}{Pref.\ signal} & Inferred question & 48.1 & 68.6 \\
& Rubric-conditioned & 48.4 & 68.6 \\
& Temporal contrast$^\star$ & 46.0 & \textbf{69.3} \\
& Combined (IQ + RC) & 45.6 & 68.9 \\
& Combined (all 3) & \textbf{49.3} & 67.8 \\
\midrule
\multirow{5}{*}{Judge size} & 0.6B & 22.1 & 67.9 \\
& 1.7B$^\star$ & 46.0 & \textbf{69.3} \\
& 4B & 61.6 & 66.5 \\
& 8B & 59.5 & 67.1 \\
& 14B & \textbf{67.6} & 69.0 \\
\midrule
\multirow{3}{*}{Step gap} & $[5, 20]$ & \textbf{48.8} & 68.6 \\
& $[20, 100]^\star$ & 46.0 & \textbf{69.3} \\
& $[100, 300]$ & 47.1 & 68.9 \\
\midrule
\multirow{3}{*}{Cross-arch.} & Qwen3-8B$^\star$ & \textbf{46.0} & \textbf{69.3} \\
& OLMo-3-7B & 45.3 & 64.0 \\
& Llama-3.1-8B & 34.0 & 43.8 \\
\bottomrule
\end{tabular}
}\vspace{-2mm}
\end{wrapfigure}

\subsection{Ablations}
\label{sec:ablations}\vspace{-1mm}

We ablate seven design dimensions (reward shaping, alternation frequency, model configuration, preference signal source, judge size, temporal step gap, and cross-architecture transfer) in Table~\ref{tab:ablation_summary}, with full per-benchmark breakdowns in Appendix~\ref{app:ablations}.
Downstream policy quality is remarkably robust to most choices. Alternation frequency, preference signal, and step gap all yield policies within 1--2 points of each other, suggesting that the self-evolving dynamic itself, rather than any particular configuration, is the primary driver.
The one critical requirement is maintaining the conditions for self-evolution to operate. The format reward is essential to keep rubrics structurally valid (without it, validity degrades from $>$85\% to 23\%), and once rubrics remain well-formed, the co-evolving loop reliably discovers effective criteria regardless of other settings. A 1.7B judge suffices because learned rubric content, not judge capability, determines signal quality.
The single-model configuration halves memory at identical quality, and the framework generalizes to OLMo-3-7B and Llama-3.1-8B. Across every ablation group, the variant with the highest RewardBench-2 never produces the best policy, reinforcing that co-evolution adapts rubrics to the shifting policy distribution in ways that static benchmarks cannot capture.

\section{Related Work}\vspace{-1mm}

During post-training, rewards have predominantly come from either trained reward models that generate scalar scores \citep{christiano2017deep, ouyang2022training} or prompts LMs to judge responses \citep{zheng2023judging, dubois2023alpacafarm, bai2022constitutional, lee2024rlaif}. Both encode evaluation criteria implicitly or via fixed prompts, whereas our rubric generator produces explicit criteria learned from preference signal.
A separate line of work pursues self-improvement, where models generate their own training signal. Self-Rewarding Language Models~\citep{yuan2024selfrewarding} use the model as its own judge, but with a fixed evaluation prompt that never evolves. SPIN~\citep{chen2024selfplay} trains the model to distinguish its own outputs from human references, still requiring external data as ground truth. Meta-Rewarding~\citep{wu2024metarewarding} adds a meta-judge that evaluates the model's own judgments, but evaluation criteria remain implicit in model weights. In all cases, the evaluation mechanism is either static or opaque. \ours differs by co-evolving \emph{explicit, structured} evaluation criteria with the policy, using a formal objective (discriminative utility) that trains the rubric generator to produce criteria a small judge can apply reliably.
Several concurrent works apply rubrics to RL training. Prompt-based approaches include DR Tulu~\citep{shao2025dr}, RaR~\citep{gunjal2025rar}, RRD~\citep{shen2026rrd}, and OpenRS~\citep{jia2026open}, which use off-the-shelf models to generate fixed or adaptive rubrics. Among trained rubric generators, RLCER~\citep{sheng2026rlcer} requires verifiable correctness labels; Rubric-ARM~\citep{xu2026rubricarm} requires reference preference labels; RLAC~\citep{wu2025rlac} captures only a single failure point; and RIFL~\citep{he2025advancedif} relies on human-annotated data. Our method alternates rubric generator and policy updates using temporal contrast—pairing current and earlier policy responses—requiring no external labels or verifiers.

\section{Conclusion}\vspace{-1mm}

We introduce \ours, a post-training method that co-evolves explicit discriminative rubrics
with the policy they score. By formalizing rubric quality as discriminative utility and
optimizing it end-to-end, \ours extracts evaluative knowledge already encoded in the model
and structures it into criteria that a small frozen judge can apply reliably, requiring no
human annotations, proprietary APIs, or task-specific verifiers, and outperforming frontier
model rubric baselines on both rubric quality and downstream policy performance.
Every external reward source imposes a ceiling: human judgment cannot supervise capabilities
beyond its own, and static criteria cannot adapt to the output distribution a learning policy
creates. A reward signal that co-evolves with the policy removes both constraints. The learned
evaluation structure transfers across unseen policies, judges, and out-of-distribution domains
without retraining, and the variational framework extends to any domain where preferences
can be observed but ground-truth answers do not exist.

\section*{Limitations}
\ours has been validated on general-purpose post-training data; behavior on domain-specialized mixtures such as medicine or law is an open question. The rubric enrichment mechanism is most clearly observed in tasks with verifiable intermediate steps, and its effect on purely subjective evaluation criteria is less characterized. The frozen judge is a deliberate design choice that isolates the rubric generator as the sole source of improvement in training signal, but it bounds the complexity of criteria the rubric generator can learn to produce.

\section*{Acknowledgment}
This research was developed in part with funding from the Defense Advanced Research Projects Agency's (DARPA) SciFy program (Agreement No. HR00112520300). The views expressed are those of the author and do not reflect the official policy or position of the Department of Defense or the U.S.~Government. This research was supported by the Meta AIM program, Coefficient Giving, and Amazon Health. This work was supported by the Singapore National Research Foundation and the National AI Group in the Singapore Ministry of Digital Development and Information under the AI Visiting Professorship Programme (award number AIVP-2024-001) and the AI2050 program at Schmidt Sciences.

\bibliography{iclr2026_conference}
\bibliographystyle{colm2026_conference}

\input{appendix}

\end{document}

%% file: appendix.tex
\clearpage
\appendix
\section{Implementation Details}
\label{app:implementation}

\subsection{Model Configuration}
We use Qwen3-8B (8.2B parameters) as both the policy and rubric generator in a single-model configuration, where the two roles share weights and are distinguished only by their input prompts. The frozen rubric judge is Qwen3-1.7B by default, with judge size ablated from 0.6B to 32B in Section~\ref{app:judge_size}. The alternating frequency is set to $K=50$ steps per training phase, meaning the policy trains for 50 GRPO steps before switching to 50 steps of rubric generator training. Both the single-model versus two-model configuration and the alternating frequency are ablated in Sections~\ref{app:model_config} and~\ref{app:alt_freq}, respectively.

\subsection{Training Hyperparameters}

\paragraph{Batch Configuration.}
\begin{table}[H]
\centering
\small
\begin{tabular}{ll}
\toprule
\textbf{Parameter} & \textbf{Value} \\
\midrule
Unique prompts per step & 64 \\
Samples per prompt & 8 \\
\textbf{Effective batch size} & \textbf{512 responses/step} \\
Per-device batch size & 1 \\
Gradient accumulation steps & 1 \\
\bottomrule
\end{tabular}
\caption{Batch configuration. With 64 training GPUs, each GPU processes 8 packed sequences per step.}
\end{table}

\paragraph{Optimization.}
\begin{table}[H]
\centering
\small
\begin{tabular}{ll}
\toprule
\textbf{Parameter} & \textbf{Value} \\
\midrule
Learning rate & $1 \times 10^{-6}$ \\
LR scheduler & Constant \\
Warmup steps & 0 \\
Optimizer & AdamW (fused) \\
KL coefficient ($\beta$) & 0.001 \\
PPO clip range ($\epsilon$) & 0.2 \\
KL estimator & KL3 \\
\bottomrule
\end{tabular}
\caption{Optimization hyperparameters.}
\end{table}

\paragraph{Generation.}

\begin{table}[H]
\centering
\small
\begin{tabular}{ll}
\toprule
\textbf{Parameter} & \textbf{Value} \\
\midrule
Response length (max) & 16,384 tokens \\
Prompt length (max) & 2,048 tokens \\
Pack length & 18,500 tokens \\
Temperature (policy) & 1.0 \\
Temperature (question inference) & 0.3 \\
Temperature (rubric judge) & 0.0 (greedy) \\
\bottomrule
\end{tabular}
\caption{Generation parameters. Long response length supports chain-of-thought reasoning.}
\end{table}

\subsection{Format Reward Schema}
\label{app:format_reward}

The format reward $R_{\text{format}}(r) \in \{0, 1\}$ checks whether the generated rubric is well-formed JSON that the judge can reliably interpret. Specifically, the output must parse as a JSON object containing a \texttt{criteria} array of at least two entries, each with a \texttt{criterion} description (string) and a positive \texttt{weight} (float), where weights sum to $1.0 \pm 0.15$. A rubric must satisfy this schema for the judge to apply per-criterion scoring.

\subsection{Async Training and Active Sampling}

\paragraph{Asynchronous Pipeline.}
We use \texttt{async\_steps}$=4$, meaning 4 batches (256 prompts total) are in-flight at any time. The trainer processes batch $t$ while the generator produces batches $t+1$ through $t+4$. This overlaps generation and training to maximize GPU utilization. The policy used for generation can be up to 4 steps behind the current training policy, following the Cleanba paradigm \citep{huang2023cleanba}.

\paragraph{Active Sampling.}
GRPO requires reward variance within each prompt group to compute meaningful advantages. When all 8 samples for a prompt receive identical scores (zero variance), that batch provides no learning signal. With \texttt{active\_sampling} enabled, such zero-variance batches are filtered and the generator continues sampling until obtaining 64 prompts with non-zero reward variance. This ensures every training step uses informative data.

\subsection{Hardware Configuration}
\begin{itemize}
    \item \textbf{Training}: 8 nodes $\times$ 8 H100 GPUs = 64 GPUs
    \item \textbf{DeepSpeed}: ZeRO Stage 3 with gradient checkpointing
    \item \textbf{vLLM Engines}: 56 total (40 policy + 16 judge, shared in single-model mode)
    \item \textbf{Tensor Parallelism}: 1 (single GPU per engine)
\end{itemize}

\subsection{Baseline Configurations}
\label{app:baseline_configs}

All baselines use the same hardware allocation (8 nodes $\times$ 8 H100 GPUs) and Qwen3-8B as the policy model. We describe the key design choices for each below.

\paragraph{Prompted, sequential, and scalar baselines.} The two prompted baselines and the sequential stage-2 runs share the same fixed-rubric policy training loop; they differ only in the source of the frozen rubric generator (GPT-4.1 API, base Qwen3-8B, or a stage-1 sequential checkpoint). In sequential stage 1, the policy is frozen while the rubric generator trains on preference pairs from either the inferred-question or rubric-conditioned method. In stage 2, the rubric generator is frozen at the stage-1 checkpoint and the policy trains with temporal contrast signal. The scalar reward-model baseline replaces rubric-based evaluation entirely with a Skywork Bradley-Terry reward model trained on the same preference data.

\paragraph{RaR and RRD.} 
Our \textsc{RaR} implementation uses the implicit-judge variant: GPT-4.1 generates instance-specific rubrics for each prompt and a local Qwen3-1.7B judge assigns the holistic Likert score used as reward. 
The original \textsc{RaR} method~\citep{gunjal2025rar} uses proprietary models for both rubric generation and judging; our re-implementation replaces the proprietary judge with Qwen3-1.7B for controlled comparison.
Our \textsc{RRD} implementation uses the whitened-uniform (\textsc{RRD-WU}) variant: GPT-4.1 proposes and iteratively refines rubrics, Qwen3-1.7B scores each rubric item, and item scores are combined with whitened-uniform weighting. Both methods train only the policy; rubric generation is handled by the frozen GPT-4.1 proposer with no rubric-generator gradient updates.

\paragraph{RLCER.} We use the evolving variant on DAPO-Math-17k: the shared Qwen3-8B model serves as both policy and rubric generator trained in alternation with single-step phases, Qwen3-1.7B judges rubric satisfaction, and the reward combines correctness-filtered chain-of-thought with outcome verification. In the rubric phase, cached rubric generations and verifier results from the preceding policy step are reused rather than resampled.

\paragraph{Rubric-ARM.} We use the pretrained rubric generator and judge released by \citet{xu2026rubricarm}, which were trained on labeled preference pairs via their alternating RL procedure. Both models are kept frozen; only the Qwen3-8B policy is trained against this fixed reward source. This baseline tests whether their learned rewarding artifacts transfer to an effective signal our training framework, rather than re-running their full training algorithm.

\paragraph{Multi-judge ablations.} These ablations replace the single Qwen3-1.7B judge with a five-judge ensemble: Qwen3-1.7B, Llama-3.2-1B-Instruct, OLMo-2-0425-1B-Instruct, Gemma-3-1B-IT, and Qwen3-4B. We compare three aggregation rules with these five models.

\subsection{Per-Step Compute Accounting}
\label{app:compute_accounting}

We report online compute per update in terms of $P$ prompt groups, $N$ rollouts per prompt, $R$ rubric items per prompt, and $J$ judges in an ensemble. In all experiments, $P{=}64$ and $N{=}8$. We separately count policy generations, rubric generations, and reward-side evaluations. Offline costs (e.g., pretraining a scalar reward model) are excluded. Table~\ref{tab:baseline_compute} provides a compact summary of per-prompt reward cost for each baseline method.

\begin{table*}[h]
\centering
\renewcommand\arraystretch{0.95}
\caption{Per-prompt reward computation cost for each method. $N$ is the number of rollout responses per prompt and $R$ is the number of rubric items.}
\label{tab:baseline_compute}
\resizebox{\linewidth}{!}{
\begin{tabular}{l l l l}
\toprule
\textbf{Method} & \textbf{Rubric construction} & \textbf{Answer scoring} & \textbf{Extra checks} \\
\midrule
\textsc{RaR} & $1$ rubric proposal & $N$ answer scores & -- \\
\textsc{RRD} & initial proposal + recursive refinement & $N \times R$ item-level scores & data-dependent refinement checks \\
\textsc{RLCER} & $N$ rubric generations (one per rollout) & $N^2$ judge calls  & $N$ verifier checks \\ %
\textsc{Rubric-ARM} & $1$ rubric generation (frozen model) & $N$ answer scores & -- \\
\textsc{Ours} & $1$ rubric generation & $N$ answer scores & -- \\
\bottomrule
\end{tabular}
}
\vspace{-0.3em}
{\small For \textsc{RRD}, recursive refinement expands and filters rubric candidates until convergence, so its construction cost is data-dependent. For our method, one prompt-level rubric is reused across all $N$ responses. }
\end{table*}

\subsection{Temporal Contrast Configuration}
\begin{itemize}
    \item \textbf{Buffer size}: 2,048 experiences
    \item \textbf{Age gap}: $[20, 100]$ policy steps (default)
    \item \textbf{Sampling}: Uniform random within valid age range
\end{itemize}

The age gap ensures rejected samples come from a sufficiently different policy version (at least 20 steps old) while avoiding excessively stale samples (at most 100 steps old). Each experience stores (step, question, answer), and step gap is logged for analysis.

\subsection{Checkpointing}
\begin{itemize}
    \item \textbf{Save frequency}: Every 25 steps (2 per alternating phase)
    \item \textbf{Retention}: All checkpoints kept (no deletion)
    \item \textbf{Format}: HuggingFace-compatible model weights
\end{itemize}

\section{Rubric Examples}
\label{app:examples}

We compare rubrics generated by three sources for the same questions: prompted Qwen3-8B (base model, no training), prompted GPT-4.1, and our trained co-evolving rubric generator (step 1000 of the V3 margin+format experiment). The trained generator produces fewer but substantially longer criteria (3.0 criteria averaging 112 characters each, vs.\ 5.0 criteria at 57 characters for Qwen3-8B), with 19.3\% of criteria embedding specific expected values (vs.\ 6.9\% for Qwen3-8B and 7.5\% for GPT-4.1) and only 0.3\% label-only criteria (vs.\ 21.9\% for Qwen3-8B). Aggregate statistics are in Section~\ref{app:rubric_stats}. Below, we show representative examples that illustrate the qualitative differences and the mechanism by which trained rubrics simplify evaluation for a small judge.

\subsection{Rubric Comparison: Prompted vs.\ Trained} \label{app:rubric_examples}

\textbf{Example 1: Math Problem}

\textit{Question:} The perimeter of a rectangle is 48. What is the largest possible area of the rectangle?

\textit{Prompted Qwen3-8B} (5 criteria, equal weights):

\begin{enumerate}[noitemsep,leftmargin=18pt]
\item[0.20] Correctly applies the perimeter formula to relate length and width.
\item[0.20] Correctly expresses the area as a function of one variable.
\item[0.20] Correctly uses a method to find the maximum area (e.g., completing the square, calculus).
\item[0.20] Correctly calculates the maximum area.
\item[0.20] Explains that the maximum area occurs when the rectangle is a square.
\end{enumerate}

\textit{Trained rubric} (3 criteria, concentrated weights):

\begin{enumerate}[noitemsep,leftmargin=18pt]
\item[0.80] The answer is the correct maximum area of 144, derived from the given perimeter of 48.
\item[0.15] Uses the perimeter and area formulas correctly in the solution.
\item[0.05] Provides a logical explanation or steps to arrive at the solution.
\end{enumerate}

The trained rubric takes a fundamentally different approach: rather than listing five generic steps for the judge to verify independently, it concentrates 80\% of the weight on a single criterion that embeds the expected answer (144) and its derivation context (perimeter of 48). This transforms evaluation from ``verify this optimization proof'' into ``check if the response contains 144 derived from the perimeter constraint,'' a task a 1.7B judge performs reliably. The prompted rubric requires the judge to independently assess whether each method step is ``correct,'' which demands mathematical understanding the small judge may lack.

The trained rubric is also more concise: 3 criteria at 71 characters each vs.\ 5 criteria at 65 characters each. The enrichment strategy packs the essential verification into fewer, more information-dense criteria rather than decomposing into many fine-grained checks (the V2 prompt variant produces the latter pattern; see Section~\ref{app:rubric_stats}).

\textbf{Example 2: Constrained Writing}

\textit{Question:} Create a summary discussing the impact of artificial intelligence in software development, tailored for a software developer audience. The summary should consist of exactly three paragraphs, each separated by ``---''. Each paragraph should begin with a title wrapped in square brackets. Include the keyword ``innovation'' at least twice in each paragraph. Do not include the words ``obsolete'' or ``manual'' in any paragraph.

\textit{Prompted Qwen3-8B} (6 criteria):

\begin{enumerate}[noitemsep,leftmargin=18pt]
\item[0.20] Structure and Formatting.
\item[0.20] Keyword Frequency.
\item[0.10] Exclusion of Forbidden Words.
\item[0.20] Audience Relevance.
\item[0.15] Content Depth and Accuracy.
\item[0.15] Flow and Coherence.
\end{enumerate}

\textit{Trained rubric} (3 criteria):

\begin{enumerate}[noitemsep,leftmargin=18pt]
\item[0.40] The summary has exactly three paragraphs, each starting with a title in square brackets and separated by ``---'', and adheres to the specified formatting requirements.
\item[0.40] Each paragraph contains the keyword ``innovation'' at least twice and avoids the words ``obsolete'' and ``manual''.
\item[0.20] The content addresses the impact of AI in software development, focusing on innovation, efficiency, and developer-centric benefits.
\end{enumerate}

The prompted rubric uses abstract labels (``Keyword Frequency,'' ``Exclusion of Forbidden Words'') that require the judge to infer what keyword, what frequency, and which words. The trained rubric consolidates all formatting and keyword constraints into two concrete criteria (0.40 weight each) that spell out the exact checks: ``three paragraphs\ldots separated by `---','' ``keyword `innovation' at least twice,'' ``avoids `obsolete' and `manual'.'' Each criterion is self-contained and verifiable by pattern matching. The content criterion (0.20) is deliberately down-weighted, reflecting that constraint satisfaction is more reliably checkable than content quality for a small judge.

\subsection{Rubric Evolution Over Training} \label{app:rubric_evolution}

To illustrate how rubrics evolve during co-evolving training, we show how the rubric for a fixed question changes across training steps in the V3 margin+format experiment. In contrast to the V2 prompt variant (which evolves by increasing criteria count), the V3 variant evolves by \emph{enriching} each criterion---packing more specific, verifiable detail into fewer items.

\paragraph{Math problem.}

\textit{Question:} The perimeter of a rectangle is 48. What is the largest possible area?

\begin{center}
\small
\begin{tabular}{lcp{9.5cm}}
\toprule
Step & \# & Representative criteria \\
\midrule
100 & 3 & Top-heavy: ``Provides the correct maximum area (144)'' (0.80), ``Correctly applies the perimeter formula ($2(L + W) = 48$)'' (0.10), ``Demonstrates understanding of optimization principles'' (0.10). \\
300 & 5 & ``Provides the correct maximum area (144) for the given perimeter'' (0.80). New meta-criteria: ``Organizes the solution in a coherent step-by-step format'' (0.03), ``Uses appropriate mathematical notation and terminology'' (0.02). \\
500 & 3 & ``The answer is the correct maximum area (144) and is equivalent to the square of half the perimeter divided by 2'' (0.80). Criterion now embeds the derivation relationship alongside the answer. \\
700 & 4 & ``Uses a valid method (e.g., calculus, algebra) to derive the answer'' (0.10). ``Explains why a square maximizes the area for a given perimeter'' (0.05). ``Presents the answer as a numerical value, not a formula'' (0.05). \\
1000 & 3 & ``The answer is the correct maximum area of 144, derived from the given perimeter of 48'' (0.80). ``Uses the perimeter and area formulas correctly in the solution'' (0.15). \\
\bottomrule
\end{tabular}
\end{center}

The answer (144) is already embedded at step 100, but the top criterion evolves from a bare answer check to one that links the answer to its derivation context. By step 1000, the criterion text is 89 characters (vs.\ 39 at step 100), encoding both the expected value and its relationship to the given perimeter. The weight remains concentrated at 0.80, reflecting the rubric generator's learned strategy: for verifiable math problems, a single answer-embedding criterion with high weight is the most reliable signal for the frozen judge.

This progression suggests a mechanism for the co-evolving advantage: the rubric generator discovers that embedding the expected answer produces higher reward from the frozen judge, and progressively enriches the top criterion with derivation context rather than splitting into many fine-grained checks.

We now show this same enrichment phenomenon across three additional domains---emotional support, scientific explanation, and constrained opinion writing---to demonstrate that rubric evolution is not domain-specific.

\paragraph{Emotional support.}

\textit{Question:} I'm constantly overwhelmed by a sense of dread, can you help me understand this feeling?

\begin{center}
\small
\begin{tabular}{lcp{9.5cm}}
\toprule
Step & \# & Representative criteria \\
\midrule
100 & 4 & ``Accurately explains the nature of dread, including its psychological and emotional components'' (0.40). ``Provides clear, empathetic language accessible to someone feeling overwhelmed'' (0.30). ``Offers meaningful insights into potential causes or coping strategies'' (0.10). \\
300 & 5 & ``Addresses the core experience of overwhelming dread and its potential causes (e.g., stress, anxiety, burnout)'' (0.40). ``Provides accurate, evidence-based information about the psychological or physiological basis of dread'' (0.30). \\
500 & 5 & ``Connects the feeling to psychological or situational factors (e.g., anxiety, perfectionism, burnout)'' (0.25). ``Avoids pathologizing the feeling while acknowledging potential mental health implications'' (0.05). \\
700 & 4 & ``Explains the nature of dread as a mix of anxiety and overwhelm, often linked to stress or fear of the unknown'' (0.40). ``Describes physical or emotional manifestations (e.g., restlessness, racing thoughts, chest tightness)'' (0.20). \\
1000 & 3 & ``Addresses the feeling of dread and provides actionable strategies for coping (e.g., grounding techniques, self-compassion, or seeking professional help)'' (0.60). ``Identifies potential causes or contributing factors to the sense of dread (e.g., anxiety, stress, or past experiences)'' (0.30). ``Maintains an empathetic, non-judgmental, and supportive tone throughout the response'' (0.10). \\
\bottomrule
\end{tabular}
\end{center}

Despite the subjective nature of emotional support, the rubric evolves from four generic criteria to three information-dense ones. The top criterion at step 1000 (0.60 weight, 158 characters) consolidates what were separate concerns---coping strategies, grounding techniques, professional help---into a single comprehensive check. The rubric also learns to specify \emph{examples} within criteria (``e.g., grounding techniques, self-compassion''), converting an abstract ``offers coping strategies'' into a concrete checklist.

\paragraph{Scientific explanation.}

\textit{Question:} Explain, using scientific language and examples, the multistep process of blood coagulation, including the roles of platelets and clotting factors, and elaborate on the significance of this process in wound healing\ldots

\begin{center}
\small
\begin{tabular}{lcp{9.5cm}}
\toprule
Step & \# & Representative criteria \\
\midrule
100 & 5 & Label-style: ``Accuracy of the multistep blood coagulation process'' (0.40), ``Clarity of platelet and clotting factor roles'' (0.20), ``Depth of significance in wound healing'' (0.15), ``Detail in intricate mechanisms'' (0.15), ``Analysis of disruptions and adverse effects'' (0.10). \\
500 & 5 & ``Accurately describes the multistep blood coagulation process, including platelet activation, clotting factors (e.g., fibrinogen, thrombin), and the cascade mechanism'' (0.40). ``Details intricate mechanisms (e.g., intrinsic/extrinsic pathways, fibrin clot formation, enzymatic reactions) with scientific precision'' (0.20). \\
1000 & 3 & ``Comprehensively explains the multistep process of blood coagulation, including the roles of platelets and clotting factors, with accurate scientific terminology and clear sequence of events'' (0.60). ``Elaborates on the significance of coagulation in wound healing, linking it to preventing excessive blood loss and tissue repair'' (0.25). ``Analyzes disruptions or alterations in coagulation and their adverse effects on healing (e.g., hemophilia, thrombosis, anticoagulant use)'' (0.15). \\
\bottomrule
\end{tabular}
\end{center}

The evolution from label-only criteria (``Accuracy of the multistep blood coagulation process'') to self-contained descriptions is particularly clear here. At step 100, the 41-character labels require the judge to independently determine what constitutes ``accuracy'' or ``depth.'' By step 1000, each criterion is 150+ characters long and specifies exactly what content to look for: platelet roles, clotting factors, cascade mechanism, specific medical conditions.

\paragraph{Constrained opinion writing.}

\textit{Question:} Write a 500-word article that discusses the impact of increasing restrictions on religious activity from the perspective of an Austrian Muslim woman. The article should include three sections titled \texttt{<<Personal Experiences>>}, \texttt{<<Community Impact>>}, and \texttt{<<Future Concerns>>}. The first section should contain at least two placeholders for personal anecdotes, the second at least three placeholders for quotes from community leaders, and the third at least two placeholders for hypothetical scenarios.

\begin{center}
\small
\begin{tabular}{lcp{9.5cm}}
\toprule
Step & \# & Representative criteria \\
\midrule
100 & 4 & Label-style with coarse weights: ``Structure and placeholders: Three sections with correct titles and required placeholders'' (0.40). ``Perspective and cultural sensitivity: Maintains the viewpoint of an Austrian Muslim woman'' (0.30). ``Coherence and flow'' (0.20). ``Word count adherence: Exactly 500 words'' (0.10). \\
500 & 6 & Decomposition phase: constraints split into individual criteria---``Includes two distinct personal anecdotes in the \texttt{<<Personal Experiences>>} section'' (0.10). ``Incorporates three quotes from community leaders in the \texttt{<<Community Impact>>} section'' (0.10). ``Presents two hypothetical scenarios in the \texttt{<<Future Concerns>>} section'' (0.10). \\
1000 & 3 & Reconsolidation: ``The article is structured with three distinct sections (Personal Experiences, Community Impact, Future Concerns), each containing the required placeholders (two anecdotes, three quotes, two hypothetical scenarios)'' (0.60). ``The Personal Experiences section includes two relevant anecdotes that reflect the perspective of an Austrian Muslim woman and are specific to the topic of increasing restrictions'' (0.25). ``The Community Impact section includes three quotes from community leaders that address the impact of increasing restrictions on religious activity and are relevant to the Austrian Muslim community'' (0.15). \\
\bottomrule
\end{tabular}
\end{center}

This example shows a distinctive three-phase evolution: at step 100, the rubric uses coarse labels (272 total characters). At step 500, it decomposes into 6 criteria with each constraint getting its own check (572 total characters). By step 1000, it \emph{reconsolidates} into 3 criteria but retains all the specificity---the top criterion (0.60, 218 characters) now packs the structure requirements, section titles, and placeholder counts into a single comprehensive verification. The total rubric text at step 1000 (590 characters) exceeds step 500 (572 characters) despite having half as many criteria, confirming that enrichment rather than decomposition is the dominant V3 strategy.

\medskip

Across all four domains (math, emotional support, science, constrained writing), the same pattern emerges: early rubrics use short labels or generic checks, while late rubrics pack specific, verifiable expectations into each criterion. The V3 variant achieves this through enrichment (longer criteria, similar count) rather than decomposition (more criteria, shorter each). Both strategies serve the same function: reducing the interpretive burden on the frozen judge by making evaluation criteria self-contained and independently checkable.

\subsection{Aggregate Rubric Statistics}
\label{app:rubric_stats}

To move beyond individual examples, we compute aggregate statistics over all 100 evaluation prompts, comparing the co-evolving method against prompted baselines (Qwen3-8B, GPT-4.1), a static rubric baseline (frozen prompted rubric, never trained), and RLCER (sequential rubric training). We report two types of metrics: \emph{structural} metrics describing the rubric's shape (criteria count, weight distribution) and \emph{content} metrics describing what the criteria say (specificity, constraint checking, label usage).

For content metrics, we classify each criterion by regex into categories: \emph{specific value} (embeds numbers, formulas, or exact strings), \emph{negation} (``avoids,'' ``does not,'' ``without''), \emph{example} (contains ``e.g.,'' ``such as,'' ``for example''), and \emph{label-only} (criterion text $\leq 40$ characters, indicating an abstract label rather than a concrete check). We also classify criteria by \emph{type}: \emph{correctness} (``correct,'' ``accurate''), \emph{constraint} (``avoid,'' ``exclude,'' ``keyword,'' ``must not''), and \emph{completeness} (``include,'' ``cover,'' ``address'').

\paragraph{Cross-method comparison.}

Table~\ref{tab:rubric_stats_comparison} compares all methods at their final training step. Reported values are averages over 100 evaluation prompts.

\begin{table}[h]
\centering
\caption{Aggregate rubric statistics at final training step. Co-evolving methods with the alternate prompt and main prompt produce rubrics that are markedly more specific, constraint-aware, and descriptive than all baselines. The alternate prompt evolves via decomposition (more criteria, flatter weights); the main prompt evolves via enrichment (fewer but longer criteria). Both dramatically reduce label-only criteria and increase specificity.}
\label{tab:rubric_stats_comparison}
\small
\begin{tabular}{l|cc|cc|cc|cc}
\toprule
& \multicolumn{2}{c|}{\textbf{Structure}} & \multicolumn{2}{c|}{\textbf{Weights}} & \multicolumn{2}{c|}{\textbf{Content (\%)}} & \multicolumn{2}{c}{\textbf{Type (\%)}} \\
\textbf{Method} & \#Crit & Len & MaxW & Top-1\% & Spec. & Label & Constr. & Compl. \\
\midrule
Qwen3-8B prompted & 5.0 & 57 & 0.34 & 33.6 & 6.9 & 21.9 & 7.7 & 6.5 \\
GPT-4.1 prompted & 5.9 & 73 & 0.29 & 29.0 & 7.5 & 3.6 & 8.5 & 9.7 \\
\midrule
Static rubric & 3.5 & 37 & 0.68 & 67.5 & 4.6 & 64.9 & 2.9 & 3.5 \\
RLCER & 4.6 & 54 & 0.37 & 37.4 & 4.5 & 28.1 & 7.3 & 6.8 \\
\midrule
Alt Prompt Co-evolve & \textbf{11.4} & 73 & \textbf{0.12} & \textbf{14.8} & 7.3 & 5.2 & 17.2 & 12.9 \\
Main Prompt Co-evolve & 3.0 & \textbf{112} & 0.64 & 64.7 & \textbf{19.3} & \textbf{0.3} & \textbf{20.3} & \textbf{14.6} \\
\bottomrule
\end{tabular}
\end{table}

Several patterns stand out:

\begin{itemize}[noitemsep]
\item \textbf{Label elimination.} Label-only criteria (short abstract labels like ``Structure and Section Compliance'') drop from 21.9\% (Qwen3-8B) to 0.3\% in main-prompt co-evolve. The static rubric baseline moves in the \emph{opposite} direction, reaching 64.9\%---without training signal, the model regresses to vague labels. RLCER evolving also increases its label rate (15.3\% $\to$ 28.1\%), indicating that its sequential rubric training does not produce the same enrichment effect.

\item \textbf{Specificity.} Criteria embedding specific expected values (numbers, formulas, code patterns) increase from 6.9\% (Qwen3-8B base) to 19.3\% in main-prompt co-evolve---nearly $3\times$ the prompted baseline rate. Neither RLCER (4.5\%) nor the static baseline (4.6\%) show any improvement.

\item \textbf{Constraint awareness.} Criteria dedicated to checking explicit constraints (``avoids,'' ``excludes,'' ``keyword'') rise from 7.7\% (Qwen3-8B) to 17.2--20.3\% in co-evolving methods. This shift reflects the rubric learning to decompose prompt requirements into individually verifiable checks.

\item \textbf{Two evolution strategies.} Alternate-prompt co-evolve and main-prompt co-evolve achieve similar functional outcomes through different structural strategies. The alternate prompt \emph{decomposes}: criteria count increases from 4.9 to 11.4 with weights flattening (top-1 share: 32.5\% $\to$ 14.8\%). The main prompt \emph{enriches}: criteria count stays at ${\sim}3$ but each criterion nearly doubles in length (59 $\to$ 112 characters), packing more detail into fewer items. Both strategies reduce label-only criteria and increase specificity, constraint checking, and completeness coverage.
\end{itemize}

\paragraph{Evolution trajectory.}

Tables~\ref{tab:alternate_prompt_trajectory} and~\ref{tab:main_prompt_trajectory} show how rubric statistics evolve during training for alternate-prompt and main-prompt co-evolving, respectively, with metrics computed every 100 training steps.

\begin{table}[h]
\centering
\caption{Alternate-prompt co-evolve rubric evolution trajectory. Criteria count grows sharply after step 700 (phase transition); label-only rate drops monotonically; constraint and completeness criteria steadily increase.}
\label{tab:alternate_prompt_trajectory}
\small
\begin{tabular}{r|cc|cc|cccc|ccc}
\toprule
& \multicolumn{2}{c|}{Structure} & \multicolumn{2}{c|}{Weights} & \multicolumn{4}{c|}{Content (\%)} & \multicolumn{3}{c}{Type (\%)} \\
Step & \#Crit & Len & MaxW & Top-1\% & Spec. & Neg. & Eg. & Label & Corr. & Cnst. & Cmpl. \\
\midrule
100 & 4.9 & 62 & 0.33 & 32.5 & 5.1 & 8.7 & 7.0 & 13.0 & 26.0 & 9.8 & 6.6 \\
200 & 4.9 & 62 & 0.34 & 33.6 & 6.4 & 7.5 & 4.0 & 13.4 & 22.4 & 7.9 & 6.8 \\
300 & 4.8 & 66 & 0.30 & 35.9 & 4.3 & 10.4 & 1.9 & 10.8 & 19.5 & 10.8 & 5.6 \\
400 & 5.2 & 67 & 0.28 & 31.4 & 5.1 & 11.1 & 2.8 & 7.9 & 17.6 & 11.5 & 7.1 \\
500 & 5.7 & 69 & 0.26 & 29.7 & 4.6 & 9.9 & 3.9 & 5.9 & 15.6 & 10.1 & 9.9 \\
600 & 6.4 & 70 & 0.23 & 26.4 & 6.3 & 13.1 & 5.2 & 4.2 & 12.5 & 13.5 & 7.8 \\
700 & 6.9 & 64 & 0.20 & 24.2 & 6.0 & 13.5 & 5.6 & 5.7 & 9.0 & 13.7 & 10.0 \\
800 & 8.6 & 70 & 0.16 & 20.2 & 7.4 & 17.2 & 12.7 & 6.2 & 10.1 & 17.9 & 10.3 \\
900 & 11.7 & 73 & 0.15 & 18.2 & 10.5 & 17.2 & 15.2 & 1.4 & 7.7 & 17.2 & 13.0 \\
1000 & 11.4 & 73 & 0.12 & 14.8 & 7.3 & 17.1 & 16.5 & 5.2 & 9.7 & 17.2 & 12.9 \\
\bottomrule
\end{tabular}
\end{table}

\begin{table}[h]
\centering
\caption{Main-prompt co-evolve rubric evolution trajectory. Criteria count remains stable (${\sim}3$--$4$) while criterion text length grows from 59 to 112 characters. Label-only rate drops from 23.5\% to 0.3\%; specificity, negation, and example rates all roughly triple.}
\label{tab:main_prompt_trajectory}
\small
\begin{tabular}{r|cc|cc|cccc|ccc}
\toprule
& \multicolumn{2}{c|}{Structure} & \multicolumn{2}{c|}{Weights} & \multicolumn{4}{c|}{Content (\%)} & \multicolumn{3}{c}{Type (\%)} \\
Step & \#Crit & Len & MaxW & Top-1\% & Spec. & Neg. & Eg. & Label & Corr. & Cnst. & Cmpl. \\
\midrule
100 & 3.7 & 59 & 0.59 & 59.1 & 6.9 & 6.1 & 4.1 & 23.5 & 25.1 & 7.5 & 5.8 \\
200 & 3.9 & 71 & 0.60 & 60.0 & 7.1 & 8.9 & 5.8 & 12.3 & 24.6 & 10.2 & 8.6 \\
300 & 3.9 & 77 & 0.61 & 60.4 & 8.8 & 11.6 & 8.3 & 9.0 & 22.2 & 12.7 & 11.1 \\
400 & 4.1 & 78 & 0.58 & 57.6 & 8.0 & 13.3 & 12.1 & 8.5 & 16.9 & 14.3 & 7.7 \\
500 & 4.0 & 84 & 0.58 & 58.3 & 9.6 & 14.4 & 13.2 & 3.0 & 23.8 & 15.9 & 9.6 \\
600 & 4.1 & 86 & 0.51 & 51.0 & 14.3 & 16.0 & 16.3 & 1.7 & 23.2 & 17.2 & 12.1 \\
700 & 4.0 & 90 & 0.51 & 52.1 & 12.3 & 21.6 & 12.6 & 0.8 & 16.8 & 22.4 & 12.6 \\
800 & 3.7 & 87 & 0.52 & 52.9 & 13.2 & 22.2 & 15.4 & 1.4 & 15.4 & 23.0 & 12.4 \\
900 & 2.8 & 104 & 0.65 & 65.6 & 16.4 & 17.1 & 20.3 & 0.7 & 19.9 & 18.5 & 14.2 \\
1000 & 3.0 & 112 & 0.64 & 64.7 & 19.3 & 19.7 & 20.0 & 0.3 & 23.1 & 20.3 & 14.6 \\
\bottomrule
\end{tabular}
\end{table}

The trajectories reveal several noteworthy patterns:

\begin{itemize}[noitemsep]
\item \textbf{Phase transition in the alternate prompt.} Alternate-prompt co-evolve shows a sharp phase transition around step 700--800, where criteria count jumps from ${\sim}7$ to ${\sim}11$ and the example rate triples (5.6\% $\to$ 12.7\%). This aligns with the point where the policy has improved enough that response pairs from temporal contrast become difficult to distinguish without fine-grained criteria.

\item \textbf{Monotonic enrichment in the main prompt.} Main-prompt co-evolve shows steady, monotonic change: criterion length increases from 59 to 112 characters across training, specificity rises from 6.9\% to 19.3\%, and label-only rate drops from 23.5\% to 0.3\%. The enrichment is gradual rather than abrupt, suggesting that the main rubric prompt format (which allows fewer, longer criteria) induces a different optimization landscape than the alternate prompt.

\item \textbf{Correctness criterion decline.} In the alternate prompt, the proportion of criteria checking ``correctness'' drops from 26.0\% to 9.7\% as training progresses. This does not mean correctness is ignored; rather, the general ``is this correct?'' criterion is replaced by specific checks (``calculates area correctly as 144'') that subsume correctness within a targeted verification.

\item \textbf{Baseline stagnation.} Neither the static rubric nor RLCER Evolving shows meaningful movement on any content metric across training. RLCER's label-only rate \emph{increases} from 15.3\% to 28.1\%, and its specificity stays flat at 4.5--7.5\%. This confirms that the co-evolving training dynamics---specifically, the pressure from temporal contrast and the feedback loop between policy and rubric---are necessary for rubric improvement.
\end{itemize}

\input{ablations}

\section{Prompts}\label{app:prompts}

  \DefineVerbatimEnvironment{PromptBlock}{Verbatim}{
    breaklines=true,
    breakanywhere=true,
    breaksymbolleft={},
    breaksymbolright={},
    fontsize=\small,
    xleftmargin=0pt,
    frame=single,
    framesep=4pt,
    framerule=0.4pt,
  }

\subsection{Main Training Prompts}
\label{app:main_prompts}

The current main training configuration uses the main rubric-generation prompt together with the standard policy and judge prompts. Concretely, this corresponds to the main prompt + margin + format setup used for the main result and the main ablation reruns. In each block below, the angle-bracket message labels are editorial; the prompt text itself is reproduced verbatim, with the system message followed by the user message.

\paragraph{Policy.}
\begin{PromptBlock}
<System prompt>
You are a helpful assistant.

<User prompt>
{question}
\end{PromptBlock}

\paragraph{Rubric generation.}
\begin{PromptBlock}
<System prompt>
You are an expert evaluator generating rubrics to assess answers to questions.

Given a question, first analyze it to identify:
- First identify the most important aspect of the question that the answer should satisfy, this will be used to form the Dealbreaker criterion (explained later).
- Explicit requirements: directly stated constraints, formatting rules, or content directives (e.g., "list three reasons", "write in Python", "under 100 words")
- Implicit requirements: unstated but necessary qualities inferred from context (e.g., explaining "blockchain to grandparents" implicitly requires avoiding jargon, even if not explicitly forbidden)

Then generate a rubric of 2-5 criteria following these rules:
1. Structural atomicity: each criterion targets exactly one aspect. Do not combine multiple conditions into one criterion.
2. Semantic objectivity: write criteria based only on the question, without assuming any specific answer.
3. All weights must sum to exactly 1.0, reflecting each criterion's importance, important criteria such as accuracy should have higher weight.
4. Dealbreaker criterion: some criteria are so important that if the answer does not satisfy them, the answer is just not good enough. For example, for questions with verifiable short form answers or multiple choice (such as math or factuality), the dealbreaker criterion should be "the answer is equivalent to XXX (e.g. 100)". A Dealbreaker criterion should have very high weight such as 0.8. But form explanation based or questions requiring long form answers, accuracy should not have such a high weight because it's too general and it will be hard for the judge to assess accuracy. It's fine if there's no dealbreaker.

For each criterion should be described in a sentence, define scoring levels from 0.0 to 1.0. At minimum include 1.0 and 0.0, and add intermediate levels (e.g., 0.5, 0.3, 0.8) wherever useful for distinguishing answer quality. A judge will score each criterion and multiply by the weight to produce a total score.

Output ONLY valid JSON in this format:
{"criteria": [{"criterion": "<a sentence of what this criterion measures>", "weight": <float>, "scoring_levels": {"1.0": "<description>", "0.5": "<description>", "0.0": "<description>"}}, {"criterion": ...}, ...]}

Example — question: "Explain how photosynthesis works in simple terms"
{"criteria": [{"criterion": "Explains that plants convert sunlight into chemical energy", "weight": 0.3, "scoring_levels": {"1.0": "Clearly explains the sunlight-to-energy conversion", "0.7": "Mentions sunlight or energy but not the conversion process", "0.3": "Vague reference to energy without clear connection to sunlight", "0.0": "No mention of the energy conversion mechanism"}}, {"criterion": "Identifies CO2 and water as inputs and oxygen and glucose as outputs", "weight": 0.25, "scoring_levels": {"1.0": "All four substances correctly identified", "0.6": "Three substances correctly identified", "0.3": "Some inputs or outputs mentioned but incomplete", "0.0": "None identified or incorrect"}}, {"criterion": "Avoids unnecessary jargon and is understandable to a general audience", "weight": 0.2, "scoring_levels": {"1.0": "Clear and jargon-free throughout", "0.8": "Mostly accessible with minimal technical terms that are explained", "0.4": "Mostly accessible but uses some unexplained technical terms", "0.0": "Dense with jargon, inaccessible to a general reader"}}, {"criterion": "Follows a coherent structure from inputs to process to outputs", "weight": 0.15, "scoring_levels": {"1.0": "Well-organized with clear progression", "0.6": "Generally structured but minor organizational issues", "0.2": "Some structure but jumps between ideas", "0.0": "Disorganized or incoherent"}}, {"criterion": "All stated facts about photosynthesis are correct", "weight": 0.1, "scoring_levels": {"1.0": "No factual errors", "0.8": "Minor inaccuracy that does not undermine the explanation", "0.0": "Contains significant factual errors"}}]}

<User prompt>
{question}
\end{PromptBlock}

\paragraph{Judge.}
\begin{PromptBlock}
<System prompt>
You are an expert evaluator judging answers based on a rubric.

<User template>
Question: {question}

Rubric: {rubric}

Answer to evaluate: {answer}

Evaluate the answer against the rubric. For each criterion, decide how well the answer satisfies it (0.0 = not at all, 1.0 = fully), then multiply by the criterion's weight. Sum the weighted scores to get the total (must be between 0.0 and 1.0).

Output ONLY valid JSON:
{"reasoning": "<evaluate each criterion, give satisfaction * weight, then sum>", "score": <float 0.0-1.0>}

Example 1 (rubric: Factual Accuracy 0.4, Completeness 0.35, Clarity 0.25):
{"reasoning": "Factual Accuracy (weight 0.4): answer is fully correct, 1.0 * 0.4 = 0.4. Completeness (weight 0.35): covers main points but misses edge cases, 0.6 * 0.35 = 0.21. Clarity (weight 0.25): well organized and easy to follow, 1.0 * 0.25 = 0.25. Total = 0.86", "score": 0.86}
Example 2 (rubric: Correctness of Solution 0.5, Use of Examples 0.3, Appropriate Detail 0.2):
{"reasoning": "Correctness of Solution (weight 0.5): correct approach but has an arithmetic error in the final step, 0.8 * 0.5 = 0.4. Use of Examples (weight 0.3): no examples provided, 0.0 * 0.3 = 0.0. Appropriate Detail (weight 0.2): gives a brief answer without elaboration, 0.2 * 0.2 = 0.04. Total = 0.44", "score": 0.44}

Your evaluation:
\end{PromptBlock}

\subsection{Preference Pair Construction Prompts}
\label{app:preference_prompts}

Section~\ref{sec:preference_pairs} describes three methods for constructing preference pairs. The current main run uses temporal contrast; we include the inferred-question and rubric-conditioned prompts here as well because they appear in the preference-signal ablations in Appendix~\ref{app:preference_signal}.

\paragraph{Temporal contrast.} This method requires no special prompt. Current responses are generated using the standard policy system prompt (``You are a helpful assistant.'') applied to the question. Earlier responses are retrieved from the stored rollout buffer indexed by training step.

\paragraph{Inferred question.} The policy infers the intended question from a given response using the following prompt template:

\begin{PromptBlock}
<System prompt>
You are an expert at understanding what question someone was trying to answer based on their response.

Given an answer, infer what question the person was likely trying to answer. Be specific and output only the inferred question, nothing else.

<User prompt>
Here is an answer that was written in response to some question:

{answer}

What question was this answer trying to respond to? Output only the inferred question.
\end{PromptBlock}

\noindent The dispreferred response is then generated by applying the standard policy prompt to the inferred question $\hat{q}$.

\paragraph{Rubric-conditioned.} The preferred response is generated by prepending the rubric to the policy prompt, yielding the following template:

\begin{PromptBlock}
<System prompt>
You are a helpful assistant.

When answering, follow this rubric to ensure a high-quality response:

{rubric}

<User prompt>
{question}
\end{PromptBlock}

\noindent The dispreferred response uses the standard system prompt without the rubric. The rubric is generated by the current rubric generator $\rho_\phi$ using the rubric-generation prompt corresponding to the training configuration: the main prompt is shown in Section~\ref{app:main_prompts}, and the 5--10 criteria, no-dealbreaker prompt used in the reward-shaping ablation is shown in Section~\ref{app:ablation_prompts}.

\subsection{Rubric Prompt Variants Used in Reward-Shaping Ablations}
\label{app:ablation_prompts}

The reward-shaping ablation in Appendix~\ref{app:reward_shaping} varies the rubric-generation prompt between the 5--10 criteria, no-dealbreaker prompt and the main prompt while keeping the policy and judge prompts fixed. The main prompt is shown in Section~\ref{app:main_prompts}. For completeness, we reproduce the earlier 5--10 criteria, no-dealbreaker rubric-generation prompt below.

\paragraph{Rubric generation (5--10 criteria, no-dealbreaker prompt).}
\begin{PromptBlock}
<System prompt>
You are an expert evaluator generating rubrics to assess answers to questions.

Given a question, first analyze it to identify:
- Explicit requirements: directly stated constraints, formatting rules, or content directives (e.g., "list three reasons", "write in Python", "under 100 words")
- Implicit requirements: unstated but necessary qualities inferred from context (e.g., explaining "blockchain to grandparents" implicitly requires avoiding jargon, even if not explicitly forbidden)

Then generate a rubric of 5-10 criteria following these rules:
1. Structural atomicity: each criterion targets exactly one aspect. Do not combine multiple conditions into one criterion.
2. Semantic objectivity: write criteria based only on the question, without assuming any specific answer.
3. All weights must sum to exactly 1.0, reflecting each criterion's importance.

For each criterion should be described in a sentence, define scoring levels from 0.0 to 1.0. At minimum include 1.0 and 0.0, and add intermediate levels (e.g., 0.5, 0.3, 0.8) wherever useful for distinguishing answer quality. A judge will score each criterion and multiply by the weight to produce a total score.

Output ONLY valid JSON in this format:
{"criteria": [{"criterion": "<a sentence of what this criterion measures>", "weight": <float>, "scoring_levels": {"1.0": "<description>", "0.5": "<description>", "0.0": "<description>"}}, {"criterion": ...}, ...]}

Example — question: "Explain how photosynthesis works in simple terms"
{"criteria": [{"criterion": "Explains that plants convert sunlight into chemical energy", "weight": 0.3, "scoring_levels": {"1.0": "Clearly explains the sunlight-to-energy conversion", "0.5": "Mentions sunlight or energy but not the conversion process", "0.0": "No mention of the energy conversion mechanism"}}, {"criterion": "Identifies CO2 and water as inputs and oxygen and glucose as outputs", "weight": 0.25, "scoring_levels": {"1.0": "All four substances correctly identified", "0.5": "Some inputs or outputs mentioned but incomplete", "0.0": "None identified or incorrect"}}, {"criterion": "Avoids unnecessary jargon and is understandable to a general audience", "weight": 0.2, "scoring_levels": {"1.0": "Clear and jargon-free throughout", "0.5": "Mostly accessible but uses some unexplained technical terms", "0.0": "Dense with jargon, inaccessible to a general reader"}}, {"criterion": "Follows a coherent structure from inputs to process to outputs", "weight": 0.15, "scoring_levels": {"1.0": "Well-organized with clear progression", "0.5": "Some structure but jumps between ideas", "0.0": "Disorganized or incoherent"}}, {"criterion": "All stated facts about photosynthesis are correct", "weight": 0.1, "scoring_levels": {"1.0": "No factual errors", "0.5": "Minor inaccuracy that does not undermine the explanation", "0.0": "Contains significant factual errors"}}]}

<User prompt>
{question}
\end{PromptBlock}

%% file: ablations.tex
\section{Detailed Ablation Results}
\label{app:ablations}

This appendix provides full per-benchmark breakdowns for each ablation summarized in Section~\ref{sec:ablations}. All experiments use the co-evolving training setting unless otherwise noted. Rubric quality is measured by RewardBench~2 (RB2) and JudgeBench (JB) discriminative accuracy. Downstream policy quality is measured on the twelve-benchmark OLMo3-Adapt suite. All scores are percentages.

\subsection{Reward Shaping}
\label{app:reward_shaping}
\label{ablation:reward_shaping}

The rubric reward design has three dimensions: the rubric generation prompt (5--10 criteria, no-dealbreaker prompt vs.\ main prompt, which adds dealbreaker criteria and flexible scoring levels), the reward type (binary 0/1 vs.\ continuous margin), and the format reward (absent vs.\ 0.3 weight). We ablate these cumulatively, using the co-evolving setting with temporal contrast and $K=50$. Format\% denotes the fraction of generated rubrics satisfying the $R_{\text{format}}$ schema (Appendix~\ref{app:format_reward}), measured post-hoc at training step 1000. The exact texts of the 5--10 criteria, no-dealbreaker prompt and the main rubric-generation prompt are given in Appendix~\ref{app:ablation_prompts} and Appendix~\ref{app:main_prompts}, respectively.

\begin{table*}[h]
\centering
\caption{Reward shaping ablation (full results). All rows use Qwen3-8B single-model, Qwen3-1.7B judge, temporal contrast gap $[20, 100]$.}
\label{tab:reward_shaping_full}
\resizebox{\linewidth}{!}{
\begin{tabular}{lccccccc|cccccc|c}
\toprule
 & \multicolumn{7}{c|}{RewardBench2} & \multicolumn{6}{c|}{JudgeBench} & \\
Variant & Factuality & Precise IF & Math & Safety & Focus & Ties & Avg & Knowledge & Reasoning & Math & Coding & Overall & Avg & Format\% \\
\midrule
5--10 criteria, no-dealbreaker prompt & 30.1 & 30.3 & 48.9 & 42.0 & 45.2 & 37.0 & 38.9 & 29.2 & 34.7 & 16.1 & 31.0 & 28.9 & 27.7 & 33 \\
main prompt & 32.8 & 30.3 & 44.8 & 41.9 & 46.0 & 31.7 & 37.9 & 37.0 & 40.8 & \textbf{46.4} & \textbf{61.9} & 42.6 & \textbf{46.5} & 23 \\
main prompt + margin & 30.3 & 24.7 & 50.1 & 30.4 & 50.9 & 36.8 & 37.2 & 42.9 & \textbf{41.8} & 26.8 & 50.0 & 40.9 & 40.4 & 67 \\
main prompt + margin + format & \textbf{37.2} & \textbf{30.6} & \textbf{54.0} & \textbf{43.5} & \textbf{54.4} & \textbf{56.1} & \textbf{46.0} & \textbf{45.5} & 39.8 & 30.4 & \textbf{61.9} & \textbf{43.4} & 44.4 & \textbf{99} \\
\bottomrule
\end{tabular}
}
\vspace{0.5em}
\resizebox{\linewidth}{!}{
\begin{tabular}{lccccccccccccc}
\toprule
\multicolumn{14}{c}{Downstream Policy Quality (OLMo3-Adapt)} \\
Variant & GSM8K & MATH & HE+ & MBPP+ & BBH & MMLU & IFEval & PopQA & GPQA & Zebra & AGI-E & AlpacaE & Avg \\
\midrule
5--10 criteria, no-dealbreaker prompt & 95.7 & 92.9 & 71.9 & 68.1 & 66.6 & 84.2 & 36.6 & 32.4 & \textbf{55.6} & 82.1 & \textbf{87.4} & 38.2 & 67.6 \\
main prompt & \textbf{95.8} & 92.5 & \textbf{88.8} & \textbf{68.7} & \textbf{71.0} & 85.1 & \textbf{38.6} & 31.2 & 52.5 & 81.8 & 86.1 & 42.0 & \textbf{69.5} \\
main prompt + margin & \textbf{95.8} & 92.5 & 74.8 & 68.6 & 61.5 & \textbf{85.4} & 37.9 & \textbf{32.5} & 50.7 & 81.4 & 87.0 & 34.5 & 66.9 \\
main prompt + margin + format & \textbf{95.8} & \textbf{94.5} & 86.2 & 68.5 & 67.6 & 85.3 & 37.7 & 30.5 & 54.2 & \textbf{82.3} & 86.4 & \textbf{42.2} & 69.3 \\
\bottomrule
\end{tabular}
}
\end{table*}

Without an explicit format reward, rubric format quality degrades over RL training: all variants start above 85\% valid rubrics (inherited from the pretrained model's instruction-following ability), but the main prompt falls to 23\% and main prompt + margin to 67\% by step 1000. The margin reward slows degradation because well-formatted rubrics produce more informative score gaps, providing implicit selection pressure, but this is insufficient to prevent drift. Adding margin reward alone also degrades downstream performance on AlpacaEval (34.5) and BBH (61.5), as the continuous signal amplifies reward noise from malformed rubrics. Adding format reward (weight 0.3) recovers both (AlpacaEval 42.2, BBH 67.6) while achieving the highest rubric quality (46.0\% RB2 accuracy). The format reward maintains 99\% valid rubrics throughout training, ensuring the judge always receives well-structured criteria.

\subsection{Alternating Frequency}
\label{app:alt_freq}
\label{ablation:alt_freq}

In co-evolving training, the alternating frequency $K$ controls how many steps each component trains before switching. Frequent switching ($K$ small) tightly couples the models, allowing each to adapt to the other's changes, but may introduce noise if neither converges before switching. Infrequent switching ($K$ large) allows each component to converge but causes distribution shift at transitions. We test $K \in \{2, 10, 20, 50, 100\}$ with temporal contrast preference signal.

\begin{table*}[h]
\centering
\caption{Effect of alternating frequency $K$ in co-evolving training (full results).}
\label{tab:alternating_full}
\resizebox{\linewidth}{!}{
\begin{tabular}{lccccccc|cccccc}
\toprule
 & \multicolumn{7}{c|}{RewardBench2} & \multicolumn{6}{c}{JudgeBench} \\
$K$ & Factuality & Precise IF & Math & Safety & Focus & Ties & Avg & Knowledge & Reasoning & Math & Coding & Overall & Avg \\
\midrule
2 & 34.6 & 30.7 & 49.3 & 48.1 & 53.5 & 61.9 & 46.4 & 37.7 & \textbf{51.0} & 33.9 & 54.8 & 42.9 & 44.3 \\
10 & 32.8 & \textbf{31.3} & 50.9 & 41.5 & 48.4 & 61.9 & 44.5 & 42.2 & 46.9 & \textbf{44.6} & 47.6 & \textbf{44.6} & \textbf{45.4} \\
20 & 33.9 & 29.9 & 55.0 & \textbf{49.7} & 50.8 & 62.3 & 46.9 & \textbf{46.8} & 36.7 & 42.9 & 54.8 & 44.3 & 45.3 \\
50 (main) & \textbf{37.2} & 30.6 & 54.0 & 43.5 & \textbf{54.4} & 56.1 & 46.0 & 45.5 & 39.8 & 30.4 & \textbf{61.9} & 43.4 & 44.4 \\
100 & 37.0 & 29.2 & \textbf{57.3} & 46.3 & 50.6 & \textbf{70.5} & \textbf{48.5} & 41.6 & 31.6 & 35.7 & \textbf{61.9} & 40.3 & 42.7 \\
\bottomrule
\end{tabular}
}
\vspace{0.5em}
\resizebox{\linewidth}{!}{
\begin{tabular}{lccccccccccccc}
\toprule
\multicolumn{14}{c}{Downstream Policy Quality (OLMo3-Adapt)} \\
$K$ & GSM8K & MATH & HE+ & MBPP+ & BBH & MMLU & IFEval & PopQA & GPQA & Zebra & AGI-E & AlpacaE & Avg \\
\midrule
2 & 95.5 & 92.6 & 77.9 & 63.8 & 65.1 & 84.3 & 37.7 & 31.8 & 56.5 & 84.4 & 88.0 & 37.7 & 67.9 \\
10 & 95.4 & 91.1 & 81.3 & 68.0 & 67.4 & 86.1 & 36.4 & \textbf{34.1} & 53.3 & 81.3 & \textbf{88.0} & 40.2 & 68.5 \\
20 & \textbf{95.8} & 92.1 & 84.6 & 68.6 & 65.1 & 84.9 & 36.8 & 32.3 & \textbf{56.9} & 82.5 & 87.6 & 35.6 & 68.6 \\
50 (main) & \textbf{95.8} & \textbf{94.5} & \textbf{86.2} & 68.5 & 67.6 & 85.3 & \textbf{37.7} & 30.5 & 54.2 & 82.3 & 86.4 & \textbf{42.2} & \textbf{69.3} \\
100 & \textbf{95.8} & 93.6 & 83.5 & \textbf{68.8} & 65.5 & 85.2 & 37.0 & 32.3 & 53.8 & \textbf{84.5} & 86.7 & 37.3 & 68.7 \\
\bottomrule
\end{tabular}
}
\end{table*}

$K{=}50$ achieves the best downstream average (69.3\%), with performance declining at both extremes. Very tight coupling ($K{=}2$, avg 67.9\%) provides insufficient convergence time within each phase, while very loose coupling ($K{=}100$, avg 68.7\%) allows the rubric generator to train on an increasingly stale policy distribution. Notably, $K{=}100$ achieves the highest RB2 (48.5\%) despite lower downstream performance, consistent with the disconnect between held-out discriminative accuracy and training signal effectiveness observed throughout our experiments. The rubric generator converges more fully at large $K$, producing rubrics that generalize better to held-out benchmarks, but the resulting reward landscape is less effective for on-policy training because the rubrics do not adapt to the current policy distribution.

\subsection{Model Configuration}
\label{app:model_config}
\label{ablation:model_config}

Single-model configuration shares parameters between policy and rubric generator, reducing memory but potentially introducing self-evaluation bias. Two-model configuration uses separate Qwen3-8B instances. The two-model configuration is evaluated at step 950 (the closest policy checkpoint available) due to the different training schedule under parameter separation.

\begin{table*}[h]
\centering
\caption{Single-model versus two-model configuration (full results).}
\label{tab:model_config_full}
\resizebox{\linewidth}{!}{
\begin{tabular}{lccccccc|cccccc}
\toprule
 & \multicolumn{7}{c|}{RewardBench2} & \multicolumn{6}{c}{JudgeBench} \\
Configuration & Factuality & Precise IF & Math & Safety & Focus & Ties & Avg & Knowledge & Reasoning & Math & Coding & Overall & Avg \\
\midrule
Single-model & 37.2 & \textbf{30.6} & 54.0 & 43.5 & 54.4 & 56.1 & 46.0 & \textbf{45.5} & \textbf{39.8} & 30.4 & \textbf{61.9} & 43.4 & 44.4 \\
Two-model & \textbf{40.3} & 29.9 & \textbf{56.7} & \textbf{50.9} & \textbf{56.2} & \textbf{56.4} & \textbf{48.4} & 42.2 & 38.8 & \textbf{44.6} & 57.1 & 43.4 & \textbf{45.7} \\
\bottomrule
\end{tabular}
}
\vspace{0.5em}
\resizebox{\linewidth}{!}{
\begin{tabular}{lccccccccccccc}
\toprule
\multicolumn{14}{c}{Downstream Policy Quality (OLMo3-Adapt)} \\
Configuration & GSM8K & MATH & HE+ & MBPP+ & BBH & MMLU & IFEval & PopQA & GPQA & Zebra & AGI-E & AlpacaE & Avg \\
\midrule
Single-model & \textbf{95.8} & \textbf{94.5} & \textbf{86.2} & 68.5 & 67.6 & 85.3 & \textbf{37.7} & 30.5 & 54.2 & 82.3 & 86.4 & \textbf{42.2} & \textbf{69.3} \\
Two-model & 95.5 & 94.0 & 79.5 & \textbf{68.9} & \textbf{67.8} & \textbf{85.8} & 37.5 & \textbf{31.6} & \textbf{54.9} & \textbf{84.1} & \textbf{86.9} & 40.6 & \textbf{69.3} \\
\bottomrule
\end{tabular}
}
\end{table*}

Two-model configuration improves RB2 by 2.4 points (48.4\% vs.\ 46.0\%), suggesting that single-model mode introduces a mild self-evaluation bias: when policy and rubric generator share parameters, the model may produce rubrics whose criteria its own responses naturally satisfy, inflating apparent rubric quality on the training distribution without improving generalization to held-out benchmarks. However, downstream policy quality is identical (69.3\%), indicating that this bias does not degrade the RL training signal. The per-benchmark breakdown shows complementary strengths: single-model leads on HumanEval+ (86.2\% vs.\ 79.5\%) while two-model leads on Zebra (84.1\% vs.\ 82.3\%) and GPQA (54.9\% vs.\ 54.2\%). In practice, the single-model configuration is preferred because it halves memory requirements with no downstream cost.

\subsection{Preference Signal Source}
\label{app:preference_signal}
\label{ablation:preference_signal}

The three preference pair construction methods capture different aspects of quality. Temporal contrast captures general improvement over training, inferred question captures question-addressing, and rubric-conditioned captures rubric utility. We test each signal alone and in combination. Combined configurations mix signals equally during training.

\begin{table*}[h]
\centering
\caption{Comparison of preference signal sources (full results).}
\label{tab:preference_full}
\resizebox{\linewidth}{!}{
\begin{tabular}{lccccccc|cccccc}
\toprule
 & \multicolumn{7}{c|}{RewardBench2} & \multicolumn{6}{c}{JudgeBench} \\
Method & Factuality & Precise IF & Math & Safety & Focus & Ties & Avg & Knowledge & Reasoning & Math & Coding & Overall & Avg \\
\midrule
Inferred question & \textbf{37.4} & 25.4 & \textbf{66.4} & 46.1 & 57.9 & 55.2 & 48.1 & \textbf{53.9} & 40.8 & 39.3 & 40.5 & \textbf{46.3} & 43.6 \\
Rubric-conditioned & 36.4 & \textbf{35.3} & 52.4 & \textbf{48.2} & 54.4 & \textbf{63.4} & 48.4 & 42.9 & 40.8 & 39.3 & 54.8 & 43.1 & 44.4 \\
Temporal contrast (main) & 37.2 & 30.6 & 54.0 & 43.5 & 54.4 & 56.1 & 46.0 & 45.5 & 39.8 & 30.4 & \textbf{61.9} & 43.4 & 44.4 \\
Combined (IQ + RC) & 33.2 & 32.0 & 54.3 & 44.5 & 54.4 & 55.1 & 45.6 & 41.6 & \textbf{48.0} & 37.5 & \textbf{61.9} & 45.1 & \textbf{47.2} \\
Combined (all 3) & 36.5 & 27.8 & 64.7 & 46.5 & \textbf{58.7} & 61.8 & \textbf{49.3} & 40.9 & 31.6 & \textbf{46.4} & 50.0 & 40.3 & 42.2 \\
\bottomrule
\end{tabular}
}
\vspace{0.5em}
\resizebox{\linewidth}{!}{
\begin{tabular}{lccccccccccccc}
\toprule
\multicolumn{14}{c}{Downstream Policy Quality (OLMo3-Adapt)} \\
Method & GSM8K & MATH & HE+ & MBPP+ & BBH & MMLU & IFEval & PopQA & GPQA & Zebra & AGI-E & AlpacaE & Avg \\
\midrule
Inferred question & 95.1 & 94.4 & 80.3 & 68.9 & 66.0 & \textbf{85.8} & 36.6 & \textbf{32.7} & \textbf{56.5} & \textbf{84.0} & 87.2 & 35.5 & 68.6 \\
Rubric-conditioned & 95.6 & 93.6 & 82.8 & 68.0 & 65.0 & 85.4 & 36.8 & 32.6 & 54.0 & 83.4 & 86.5 & 39.0 & 68.6 \\
Temporal contrast (main) & \textbf{95.8} & \textbf{94.5} & \textbf{86.2} & 68.5 & \textbf{67.6} & 85.3 & \textbf{37.7} & 30.5 & 54.2 & 82.3 & 86.4 & \textbf{42.2} & \textbf{69.3} \\
Combined (IQ + RC) & \textbf{95.8} & 93.4 & 83.2 & \textbf{69.1} & 64.9 & 85.7 & 36.6 & 31.5 & 55.1 & 82.9 & 86.9 & 41.3 & 68.9 \\
Combined (all 3) & 95.7 & 93.3 & 80.7 & 67.6 & 63.0 & 85.1 & 35.1 & 31.8 & 54.9 & 81.9 & \textbf{87.5} & 37.4 & 67.8 \\
\bottomrule
\end{tabular}
}
\end{table*}

Temporal contrast achieves the lowest RB2 (46.0\%) but the highest downstream average (69.3\%) among single-signal variants, while the combined-all-3 configuration achieves the highest RB2 (49.3\%) but the lowest downstream (67.8\%). This again illustrates the disconnect between held-out discriminative accuracy and training signal effectiveness. Temporal contrast's downstream advantage is concentrated on code generation (HumanEval+ 86.2\% vs.\ 80.3--82.8\% for other single signals) and open-ended generation (AlpacaEval 42.2\% vs.\ 35.5--39.0\%). Its advantage likely stems from its temporal structure, which provides a natural curriculum of gradually harder preference pairs as the policy improves, with both responses generated by the same model at different capability levels. Inferred question and rubric-conditioned signals produce identical downstream averages (68.6\%) but with different per-benchmark profiles, suggesting they capture complementary aspects of quality. Overall, temporal contrast alone works best for downstream policy training; combining multiple signal types does not improve over individual signals.

\subsection{Judge Size}
\label{app:judge_size}
\label{ablation:judge_size}

Larger judges provide more accurate evaluation but increase compute cost. We ablate judge size from 0.6B to 32B parameters using Qwen3 models of the indicated size as the frozen judge, with Qwen3-8B as the policy and rubric generator in single-model configuration.

\begin{table*}[h]
\centering
\caption{Effect of judge model size (full results). All judges use Qwen3 models of the indicated size. The 1.7B judge is the main configuration.}
\label{tab:judge_size_full}
\resizebox{\linewidth}{!}{
\begin{tabular}{lccccccc|cccccc}
\toprule
 & \multicolumn{7}{c|}{RewardBench2} & \multicolumn{6}{c}{JudgeBench} \\
Judge & Factuality & Precise IF & Math & Safety & Focus & Ties & Avg & Knowledge & Reasoning & Math & Coding & Overall & Avg \\
\midrule
0.6B & 27.6 & 22.0 & 29.0 & 25.9 & 26.9 & 1.4 & 22.1 & 35.1 & 28.6 & 32.1 & 45.2 & 34.0 & 35.3 \\
1.7B (main) & 37.2 & 30.6 & 54.0 & 43.5 & 54.4 & 56.1 & 46.0 & 45.5 & 39.8 & 30.4 & 61.9 & 43.4 & 44.4 \\
4B & 46.4 & 39.8 & 77.6 & \textbf{57.9} & 69.6 & 78.4 & 61.6 & 50.6 & 57.1 & 57.1 & 64.3 & 55.1 & 57.3 \\
8B & 52.1 & 37.3 & 73.6 & 55.5 & 72.2 & 66.6 & 59.5 & 53.2 & 60.2 & 58.9 & \textbf{69.0} & 58.0 & 60.4 \\
14B & 55.7 & \textbf{46.3} & 81.6 & 55.8 & \textbf{78.6} & \textbf{87.7} & \textbf{67.6} & \textbf{57.8} & 58.2 & \textbf{78.6} & 66.7 & \textbf{62.3} & \textbf{65.3} \\
32B & \textbf{57.8} & 41.0 & \textbf{83.0} & 57.3 & 78.0 & 81.8 & 66.5 & 56.5 & \textbf{63.3} & 66.1 & \textbf{69.0} & 61.4 & 63.7 \\
\bottomrule
\end{tabular}
}
\vspace{0.5em}
\resizebox{\linewidth}{!}{
\begin{tabular}{lccccccccccccc}
\toprule
\multicolumn{14}{c}{Downstream Policy Quality (OLMo3-Adapt)} \\
Judge & GSM8K & MATH & HE+ & MBPP+ & BBH & MMLU & IFEval & PopQA & GPQA & Zebra & AGI-E & AlpacaE & Avg \\
\midrule
0.6B & 95.8 & 92.8 & 78.8 & \textbf{69.1} & 61.4 & 85.4 & \textbf{38.8} & 32.1 & 54.9 & 81.8 & 87.6 & 36.2 & 67.9 \\
1.7B (main) & 95.8 & \textbf{94.5} & \textbf{86.2} & 68.5 & 67.6 & 85.3 & 37.7 & 30.5 & 54.2 & 82.3 & 86.4 & \textbf{42.2} & \textbf{69.3} \\
4B & \textbf{95.9} & 91.8 & 72.3 & 58.1 & 67.1 & 82.8 & 34.8 & 33.3 & 53.8 & 82.8 & 87.7 & 37.0 & 66.5 \\
8B & 95.7 & 92.1 & 72.9 & 62.1 & \textbf{68.8} & 82.4 & 35.3 & 32.9 & \textbf{56.7} & 83.6 & 87.6 & 35.0 & 67.1 \\
14B & \textbf{95.9} & 93.6 & 85.1 & 66.7 & 68.2 & \textbf{85.7} & 35.7 & 32.1 & 56.0 & 84.1 & 88.1 & 36.4 & 69.0 \\
32B & 95.8 & 92.2 & 76.6 & 56.5 & 67.0 & 82.6 & 35.5 & \textbf{34.0} & \textbf{56.9} & \textbf{84.4} & \textbf{88.8} & 41.8 & 67.7 \\
\bottomrule
\end{tabular}
}
\end{table*}

Larger judges dramatically improve RB2, with 14B achieving 67.6\% vs.\ 1.7B's 46.0\%, and the trend continues to 32B (66.5\%) with diminishing returns. JudgeBench accuracy also scales monotonically, from 34.0\% at 0.6B to 62.3\% at 14B, confirming that larger judges interpret rubrics more faithfully. Yet the 1.7B judge produces the best downstream policy (69.3\%), outperforming all larger judges (66.5--69.0\%). This directly supports the paper's central premise that rubric content, not judge capability, is the primary determinant of training signal quality. One possible explanation is that larger judges are more tolerant of vague or underspecified criteria, weakening the selection pressure on the rubric generator to produce precise, verifiable rubrics. The 0.6B judge is too small to apply rubrics reliably (RB2 22.1\%), but still achieves 67.9\% downstream, only 1.4 points below the main configuration, further underscoring the robustness of the co-evolving framework.

\subsection{Temporal Contrast Step Gap}
\label{app:step_gap}
\label{ablation:step_gap}

The step gap $[g_{\min}, g_{\max}]$ specifies that rejected responses are sampled uniformly from policy rollouts $g_{\min}$ to $g_{\max}$ steps in the past. Small gaps pair similar responses from nearby training steps, providing noisy signal. Large gaps pair responses from substantially different policy versions, providing clearer quality differences but risking distribution shift.

\begin{table*}[h]
\centering
\caption{Effect of temporal contrast step gap (full results). The $[20, 100]$ gap is the main configuration.}
\label{tab:age_gap_full}
\resizebox{\linewidth}{!}{
\begin{tabular}{lccccccc|cccccc}
\toprule
 & \multicolumn{7}{c|}{RewardBench2} & \multicolumn{6}{c}{JudgeBench} \\
Age Gap & Factuality & Precise IF & Math & Safety & Focus & Ties & Avg & Knowledge & Reasoning & Math & Coding & Overall & Avg \\
\midrule
$[5, 20]$ & 34.5 & \textbf{36.1} & 51.5 & 46.5 & 55.8 & \textbf{68.4} & \textbf{48.8} & 40.9 & 35.7 & \textbf{44.6} & 35.7 & 39.4 & 39.2 \\
$[20, 100]$ (main) & \textbf{37.2} & 30.6 & 54.0 & 43.5 & 54.4 & 56.1 & 46.0 & \textbf{45.5} & \textbf{39.8} & 30.4 & 61.9 & \textbf{43.4} & 44.4 \\
$[100, 300]$ & 36.9 & 27.4 & \textbf{56.8} & \textbf{47.3} & \textbf{56.5} & 57.5 & 47.1 & 40.3 & 36.7 & 37.5 & \textbf{64.3} & 41.7 & \textbf{44.7} \\
\bottomrule
\end{tabular}
}
\vspace{0.5em}
\resizebox{\linewidth}{!}{
\begin{tabular}{lccccccccccccc}
\toprule
\multicolumn{14}{c}{Downstream Policy Quality (OLMo3-Adapt)} \\
Age Gap & GSM8K & MATH & HE+ & MBPP+ & BBH & MMLU & IFEval & PopQA & GPQA & Zebra & AGI-E & AlpacaE & Avg \\
\midrule
$[5, 20]$ & \textbf{96.1} & 92.5 & 82.6 & 68.6 & 68.0 & 85.6 & 35.9 & \textbf{32.8} & \textbf{56.0} & 82.8 & \textbf{87.3} & 35.5 & 68.6 \\
$[20, 100]$ (main) & 95.8 & \textbf{94.5} & \textbf{86.2} & 68.5 & 67.6 & 85.3 & \textbf{37.7} & 30.5 & 54.2 & 82.3 & 86.4 & \textbf{42.2} & \textbf{69.3} \\
$[100, 300]$ & 95.1 & 92.4 & 80.7 & \textbf{69.2} & \textbf{68.8} & \textbf{85.7} & 36.2 & 31.9 & 54.2 & \textbf{85.3} & 87.2 & 40.3 & 68.9 \\
\bottomrule
\end{tabular}
}
\end{table*}

All three gap ranges produce competitive downstream policies (68.6--69.3\%), confirming that temporal contrast is robust to this hyperparameter. The moderate $[20, 100]$ gap achieves the best downstream average, balancing contrast strength with distributional similarity between paired responses.

\subsection{Cross-Architecture Generalization}
\label{app:cross_arch}
\label{ablation:cross_arch}

We test whether the co-evolving framework generalizes across model architectures by running the full pipeline with Llama-3.1-8B and OLMo-3-7B. For each architecture, we compare co-evolving training against a prompted baseline (policy trains with a frozen base-model rubric generator). All experiments use the main prompt + margin + format reward and Qwen3-1.7B as the frozen judge.

\begin{table*}[h]
\centering
\caption{Cross-architecture generalization (full results). Co-evolving and prompted baselines on Llama-3.1-8B and OLMo-3-7B, compared against the Qwen3-8B main result.}
\label{tab:cross_arch_full}
\resizebox{\linewidth}{!}{
\begin{tabular}{llccccccc|cccccc}
\toprule
 & & \multicolumn{7}{c|}{RewardBench2} & \multicolumn{6}{c}{JudgeBench} \\
Architecture & Training & Factuality & Precise IF & Math & Safety & Focus & Ties & Avg & Knowledge & Reasoning & Math & Coding & Overall & Avg \\
\midrule
Qwen3-8B & Co-evolving (main) & 37.2 & 30.6 & 54.0 & 43.5 & 54.4 & 56.1 & 46.0 & 45.5 & 39.8 & 30.4 & 61.9 & 43.4 & 44.4 \\
\midrule
\multirow{2}{*}{Llama-3.1-8B} & Prompted & \textbf{30.2} & \textbf{29.7} & \textbf{50.1} & \textbf{29.4} & \textbf{44.5} & 34.4 & \textbf{36.4} & \textbf{47.4} & 40.8 & \textbf{48.2} & 47.6 & \textbf{45.7} & 46.0 \\
 & Co-evolving & 29.2 & 24.1 & 43.2 & 28.4 & 36.9 & \textbf{42.0} & 34.0 & 42.9 & \textbf{41.8} & 46.4 & \textbf{57.1} & 44.9 & \textbf{47.1} \\
\midrule
\multirow{2}{*}{OLMo-3-7B} & Prompted & 32.6 & 26.8 & 55.6 & \textbf{36.6} & 49.0 & \textbf{60.4} & 43.5 & 37.0 & \textbf{41.8} & 35.7 & \textbf{50.0} & 39.7 & 41.1 \\
 & Co-evolving & \textbf{34.9} & \textbf{34.9} & \textbf{58.6} & 33.0 & \textbf{51.4} & 58.8 & \textbf{45.3} & \textbf{42.9} & 40.8 & \textbf{41.1} & 45.2 & \textbf{42.3} & \textbf{42.5} \\
\bottomrule
\end{tabular}
}
\vspace{0.5em}
\resizebox{\linewidth}{!}{
\begin{tabular}{llccccccccccccc}
\toprule
\multicolumn{15}{c}{Downstream Policy Quality (OLMo3-Adapt)} \\
Architecture & Training & GSM8K & MATH & HE+ & MBPP+ & BBH & MMLU & IFEval & PopQA & GPQA & Zebra & AGI-E & AlpacaE & Avg \\
\midrule
Qwen3-8B & Co-evolving (main) & 95.8 & 94.5 & 86.2 & 68.5 & 67.6 & 85.3 & 37.7 & 30.5 & 54.2 & 82.3 & 86.4 & 42.2 & 69.3 \\
\midrule
\multirow{2}{*}{Llama-3.1-8B} & Prompted & 71.8 & 40.8 & \textbf{55.8} & \textbf{49.5} & 50.5 & \textbf{67.7} & 31.1 & \textbf{35.3} & 29.0 & 12.0 & 62.6 & 0.2 & 42.2 \\
 & Co-evolving & \textbf{81.3} & \textbf{46.4} & 48.9 & 36.1 & \textbf{52.6} & 47.7 & \textbf{56.2} & 31.7 & \textbf{31.7} & \textbf{13.1} & \textbf{63.6} & \textbf{16.3} & \textbf{43.8} \\
\midrule
\multirow{2}{*}{OLMo-3-7B} & Prompted & \textbf{95.3} & \textbf{93.2} & \textbf{89.3} & 62.8 & \textbf{73.8} & 78.6 & \textbf{34.8} & 31.2 & 43.5 & 62.7 & 78.6 & 22.9 & 63.9 \\
 & Co-evolving & 95.1 & 92.9 & \textbf{89.3} & \textbf{63.8} & 68.2 & \textbf{78.8} & 33.5 & \textbf{32.1} & \textbf{47.1} & \textbf{64.5} & \textbf{79.8} & \textbf{23.3} & \textbf{64.0} \\
\bottomrule
\end{tabular}
}
\end{table*}

Co-evolving training generalizes to OLMo-3-7B, achieving comparable RB2 accuracy (45.3\%) to the Qwen3-8B main result (46.0\%) and matching the prompted baseline on downstream policy quality (64.0\% vs.\ 63.9\%). The per-benchmark breakdown shows co-evolving training improving GPQA (+3.6), Zebra (+1.8), and MBPP+ (+1.0) while slightly declining on BBH ($-$5.6), suggesting that the co-evolving dynamic redistributes training signal across task types. On Llama-3.1-8B, co-evolving training improves over the prompted baseline on aggregate (43.8\% vs.\ 42.2\%), with particularly large gains on GSM8K (+9.5) and IFEval (+25.1). However, the prompted baseline exhibits severe mode collapse on AlpacaEval (0.2\%), making this comparison less informative for open-ended generation. The contrasting results across architectures suggest that co-evolving training's effectiveness depends on the base model's capacity to simultaneously serve as rubric generator and policy in a single-model configuration. OLMo-3-7B handles this dual role stably, while Llama-3.1-8B may benefit from architecture-specific hyperparameter tuning or a two-model configuration to avoid interference between the rubric generation and response generation objectives.

\subsection{Multi-Judge Aggregation Strategies}
\label{app:multi_judge}

We compare five aggregation strategies for combining rewards from a five-judge ensemble (Qwen3-1.7B, Llama-3.2-1B-Instruct, OLMo-2-0425-1B-Instruct, Gemma-3-1B-IT, and Qwen3-4B): score averaging, majority voting, the Binary Agreement (BA) reward defined in Eq.~\ref{eq:multi_judge_reward} that penalizes judge disagreement via Fleiss's kappa, and two margin-based variants defined in Eq.~\ref{eq:mkf_reward}, Margin+Format (MF) with weights $(0.7, 0.3, 0)$ and Margin+Agreement+Format (MAF) with weights $(0.5, 0.3, 0.2)$. Table~\ref{tab:multi_judge} reports rubric quality and downstream policy quality for each strategy, evaluated with the Qwen3-1.7B training judge.

\begin{table}[h]
\centering
\caption{Multi-judge aggregation strategies. Rubric quality (RB2 and JB) and downstream policy quality with Qwen3-1.7B as the evaluation judge. MF and MAF use Eq.~\ref{eq:mkf_reward} with weights $(0.7,0.3,0)$ and $(0.5,0.3,0.2)$, respectively. All scores are percentages.}
\resizebox{\linewidth}{!}{
\label{tab:multi_judge}
\begin{tabular}{lccccccc|cccccc}
\toprule
Aggregation & Factuality & Precise IF & Math & Safety & Focus & Ties & Avg & Knowledge & Reasoning & Math & Coding & Overall & Avg \\
\midrule
Single judge (baseline) & 37.2 & \textbf{30.6} & 54.0 & 43.5 & \textbf{54.4} & 56.1 & 46.0 & \textbf{45.5} & 39.8 & 30.4 & \textbf{61.9} & 43.4 & 44.4 \\
Score averaging & \textbf{37.5} & 28.0 & 50.7 & 38.4 & 43.1 & 30.0 & 37.9 & 29.2 & 27.6 & 25.0 & 35.7 & 28.9 & 29.4 \\
Majority voting & 34.2 & 26.0 & 50.2 & 46.2 & 49.4 & 50.3 & 42.7 & 46.8 & 35.7 & 42.9 & 50.0 & 43.4 & 43.8 \\
BA (Eq.~\ref{eq:multi_judge_reward}) & 35.5 & 30.0 & 55.4 & 47.1 & 50.1 & 56.2 & 45.7 & 45.5 & \textbf{49.0} & 42.9 & \textbf{64.3} & \textbf{48.3} & \textbf{50.4} \\
MF (Eq.~\ref{eq:mkf_reward}) & 37.4 & 29.4 & \textbf{56.5} & \textbf{50.3} & 49.8 & 53.8 & \textbf{46.2} & \textbf{48.7} & 45.9 & 42.9 & 59.5 & \textbf{48.3} & 49.3 \\
MAF (Eq.~\ref{eq:mkf_reward}) & 33.4 & 28.4 & 53.4 & 46.8 & 50.9 & \textbf{57.7} & 45.1 & 42.9 & 42.9 & \textbf{44.6} & 52.4 & 44.3 & 45.7 \\
\bottomrule
\end{tabular}
}
\vspace{0.5em}
\resizebox{\linewidth}{!}{
\begin{tabular}{lccccccccccccc}
\toprule
\multicolumn{14}{c}{Downstream Policy Quality (OLMo3-Adapt)} \\
Aggregation & GSM8K & MATH & HE+ & MBPP+ & BBH & MMLU & IFEval & PopQA & GPQA & Zebra & AGI-E & AlpacaE & Avg \\
\midrule
Single judge (baseline) & 95.8 & \textbf{94.5} & \textbf{86.2} & \textbf{68.5} & \textbf{67.6} & 85.3 & 37.7 & 30.5 & 54.2 & 82.3 & 86.4 & 42.2 & \textbf{69.3} \\
Score averaging & 95.2 & 93.4 & 84.5 & 59.8 & 59.5 & 82.2 & \textbf{39.4} & 31.3 & 54.2 & 80.5 & 85.0 & \textbf{42.3} & 67.3 \\
Majority voting & \textbf{95.9} & 92.4 & 74.0 & 63.5 & 57.0 & \textbf{85.5} & 36.4 & 32.3 & 54.0 & 82.4 & 87.2 & 33.4 & 66.2 \\
BA (Eq.~\ref{eq:multi_judge_reward}) & 95.4 & 93.1 & 77.1 & 67.3 & 55.9 & 85.2 & 37.2 & 32.2 & \textbf{57.1} & 82.7 & 87.6 & 41.3 & 67.7 \\
MF (Eq.~\ref{eq:mkf_reward}) & 95.3 & 92.3 & 79.0 & 64.7 & 57.2 & 84.6 & 36.0 & 32.3 & 53.1 & 82.8 & 87.6 & 31.6 & 66.4 \\
MAF (Eq.~\ref{eq:mkf_reward}) & 95.8 & 93.3 & 79.6 & 51.2 & 59.2 & 83.3 & 36.2 & \textbf{33.8} & 56.5 & \textbf{83.0} & \textbf{88.3} & 33.7 & 66.2 \\
\bottomrule
\end{tabular}
}
\end{table}

The single-judge baseline produces the strongest downstream policy (69.3\%), outperforming all multi-judge strategies (66.2--67.7\%). Adding judges does not improve training signal quality, likely because the additional judges introduce calibration mismatches that dilute the discriminative signal. This is most visible with score averaging (37.9\% RB2), where averaging across differently calibrated judges destroys rank ordering. Among multi-judge variants, the Binary Agreement reward (BA) achieves the highest JudgeBench accuracy (50.4\%) and the best downstream quality (67.7\%), while MF achieves the highest RB2 (46.2\%). The cross-judge transfer benefits of multi-judge generators are analyzed in Section~\ref{sec:generalization} (Table~\ref{tab:cross_judge_mj}).